\documentclass[journal]{IEEEtran}
\usepackage{cite}
\usepackage{array}
\usepackage{amsmath,amssymb,amsfonts}
\usepackage{algorithmic}
\usepackage[linesnumbered,ruled,vlined]{algorithm2e}
\usepackage{graphicx}
\usepackage{textcomp}
\usepackage{xcolor}
\usepackage{enumerate}
\usepackage{booktabs}
\usepackage{diagbox}
\usepackage{array}
\usepackage{supertabular}
\usepackage{caption}
\usepackage{float} 
\usepackage{subcaption}
\usepackage{verbatim}
\usepackage{bm}
\usepackage{pifont}
\usepackage{stfloats}
\usepackage{amsthm}
\usepackage{hyperref}
\usepackage{enumitem}
\usepackage[font=footnotesize,labelfont=footnotesize]{caption}

\begin{document}

\title{Confidence-Regulated Generative Diffusion Models for Reliable AI Agent Migration in Vehicular Metaverses}
	
\author{Yingkai Kang, Jiawen Kang, Jinbo Wen, Tao Zhang, Zhaohui Yang, \\Dusit Niyato, \textit{Fellow, IEEE}, and Yan Zhang, \textit{Fellow, IEEE}

\thanks{        
        Y. Kang and J. Kang are with the School of Automation, Guangdong University of Technology, Guangzhou 510006, China (e-mails: 3122000883@mail2.gdut.edu.cn; kavinkang@gdut.edu.cn).
        
        J. Wen is with the College of Computer Science and Technology, Nanjing University of Aeronautics and Astronautics, Nanjing 210016, China (e-mail: jinbo1608@nuaa.edu.cn). 

        T. Zhang is with the School of Cyberspace Science and Technology, Beijing Jiaotong University, Beijing 100044, China (e-mail: taozh@bjtu.edu.cn).

        Z. Zhang is with the College of Information Science and Electronic Engineering, Zhejiang Provincial Key Lab of information processing, communication and networking, Zhejiang University, Hangzhou 310007, China (e-mail: yang\_zhaohui@zju.edu.cn).

        D. Niyato is with the College of Computing and Data Science, Nanyang Technological University, Singapore (e-mail: dniyato@ntu.edu.sg).

        Y. Zhang is with the Department of Informatics, University of Oslo, Norway, and also with the Simula Research Laboratory, Norway (e-mail: yanzhang@ieee.org). 

        \textit{Corresponding author: Jiawen Kang.}
	} 
}
	
\maketitle

\begin{abstract}
Vehicular metaverses are an emerging paradigm that merges intelligent transportation systems with virtual spaces, leveraging advanced digital twin and Artificial Intelligence (AI) technologies to seamlessly integrate vehicles, users, and digital environments. In this paradigm, vehicular AI agents are endowed with environment perception, decision-making, and action execution capabilities, enabling real-time processing and analysis of multi-modal data to provide users with customized interactive services. Since vehicular AI agents require substantial resources for real-time decision-making, given vehicle mobility and network dynamics conditions, the AI agents are deployed in RoadSide Units (RSUs) with sufficient resources and dynamically migrated among them. However, AI agent migration requires frequent data exchanges, which may expose vehicular metaverses to potential cyber attacks. To this end, we propose a reliable vehicular AI agent migration framework, achieving reliable dynamic migration and efficient resource scheduling through cooperation between vehicles and RSUs. Additionally, we design a trust evaluation model based on the theory of planned behavior to dynamically quantify the reputation of RSUs, thereby better accommodating the personalized trust preferences of users. We then model the vehicular AI agent migration process as a partially observable markov decision process and develop a Confidence-regulated Generative Diffusion Model (CGDM) to efficiently generate AI agent migration decisions. Numerical results demonstrate that the CGDM algorithm significantly outperforms baseline methods in reducing system latency and enhancing robustness against cyber attacks.

\end{abstract}

\IEEEpeerreviewmaketitle

\section{Introduction}

Metaverses are expected to build a unified network that seamlessly integrates physical and virtual spaces~\cite{li2022internet}. The rapid development of key technologies, including Extended Reality (XR), Digital Twin (DT), and generative Artificial Intelligence (AI), has provided a robust technical foundation for enabling immersive real-time interactions~\cite{xu2022full}. Vehicular metaverses are emerging as a new paradigm that integrates Intelligent Transportation Systems (ITSs) with metaverses~\cite{xu2023generative}. By incorporating technologies such as Augmented Virtual (AR) head-up displays, DT systems, and personalized content recommendations, the vehicular metaverse expands real-time virtual interactions of vehicles in the physical environment, providing users with immersive experiences that go beyond traditional ITSs. In vehicular metaverses, AI agents are regarded as key enablers~\cite{chamola2024beyond,xu2023generative1}. Vehicles are mapped to vehicular AI agents, comprising perception, brain, and action modules. These AI agents effectively integrate and reason about multi-modal data through Large Language Models (LLMs) and Visual Language Models (VLMs)~\cite{xi2025rise,hong2024cogagent}, providing users with immersive metaverse services. Specifically, vehicular AI agents continuously collect multi-modal data from in-vehicle and external environments through onboard sensors and perception modules. The collected data is converted by the perception module into a format that LLMs and VLMs can understand and then is conveyed to the brain module, where comprehensive decisions are generated through reasoning, planning, and memory~\cite{xi2025rise}. Subsequently, the action module executes these decisions, thereby providing users with immersive services (e.g., AR navigation and 3D virtual meeting rooms)~\cite{xi2025rise}.

Due to the requirement for vehicular AI agents to collect multi-modal data and perform reasoning in real-time, the resulting computation-intensive tasks require substantial computing resources. The local computing power of vehicles is insufficient to meet these real-time processing requirements~\cite{li2022internet}. Consequently, deploying AI agents on RoadSide Units (RSUs) with large-scale computing resources has been proposed as a promising solution~\cite{kang2024uav,xu2023generative1}. Given that the vehicular network environment is constantly changing as vehicles move at high speeds while the load and channel conditions of RSUs are highly dynamic~\cite{chen2023multiagent}, AI agents need to be migrated among RSUs in real time to maintain vehicular service continuity and low latency. Moreover, each RSU typically needs to handle multiple vehicle requests concurrently. Therefore, it is essential to optimize resource allocation and task scheduling for AI agent migration.

During the AI agent migration process, frequent data exchanges between vehicles and RSUs make vehicular metaverses highly susceptible to cyber attacks. For example, attackers may launch Distributed Denial of Service (DDoS) attacks against specific RSUs, thereby depleting their communication and computing resources~\cite{zhang2023moving}. This can disrupt normal data transmission and task scheduling at the targeted RSUs, ultimately exposing vehicular AI agent migration to significant latency or even failure. The Moving Target Defense (MTD) strategy is proposed to reduce the probability of successful attacks. By randomizing network topology and reconfiguring addresses, this defense forces the attacker to repeat reconnaissance, thereby shortening the attack window and raising the attack cost~\cite{9462504}. However, the MTD strategy cannot cover all attack windows, and RSUs remain exposed to security risks~\cite{9047923}. To identify compromised RSUs, trust mechanisms have received significant attention~\cite{huang2023trust,kang2024hybrid,ismail2024enhancing,zhong2023blockchain}, which quantify the credibility of RSUs through trust evaluation models. The widely used subjective logic model dynamically updates RSU reputation values based on user feedback but fails to fully consider interference from malicious evaluators~\cite{zhong2023blockchain}. Some recent studies have proposed multi‑level trust evaluation strategies that integrate multi‑source trust evidence to enhance the reliability of trust evaluations~\cite{kang2024hybrid,huang2023trust}. Although these methods consider both objective metrics and subjective feedback, they are still insufficient in capturing user preferences and environment differences.

To address the above challenges, we propose a reliable vehicular AI agent migration framework in vehicular metaverses. Specifically, we design a trust evaluation model based on the Theory of Planned Behavior (TPB) to dynamically compute the reputation values of RSUs. During the vehicular AI agent migration, both vehicles and RSUs jointly formulate decisions based on vehicle mobility, task requirements, and the reputation values of RSUs, ensuring low-latency and high-reliability services in a dynamic network. Given that the AI agent migration problem is NP-hard~\cite{kang2024uav}, traditional optimization algorithms (e.g., heuristic algorithms) are insufficient to meet the real-time requirements in migration scenarios due to their limited efficiency. Fortunately, Generative Diffusion Models (GDMs) for Deep Reinforcement Learning (DRL) have shown great potential in network optimization~\cite{du2024enhancing,du2024diffusion,du2023exploring}. However, most diffusion policies in DRL face policy update instability due to the high variance of policy gradients~\cite{psenka2024qsm}. To overcome this, we innovatively propose a Confidence-regulated GDM (CGDM) algorithm that enhances the final convergence performance while ensuring stable policy updates. 

Our main contributions are summarized as follows:
\begin{itemize}
    \item We design a highly reliable AI agent migration framework in vehicular metaverses, ensuring the low-latency and high-reliability requirements of AI agent migration. In this framework, we propose a TPB-based trust evaluation model to ensure the security of vehicular AI agent migration while meeting diverse user trust preferences. Moreover, we introduce bilateral collaboration between vehicles and RSUs in AI agent migration for the first time, which has not been considered in previous work.
    \item In the TPB-based trust evaluation model, by quantifying user attitudes, subjective norms, and perceived behavioral control, we can model the multi-dimensional behavioral intentions underlying the selection of RSUs, thereby providing a basis for AI agents to formulate reliable and personalized AI agent migration decisions in complex vehicular network environments.
    \item To achieve high reliability and low latency in vehicular AI agent migration, we formulate the task as a minimum‑latency optimization problem subject to resource and security constraints. We model this problem as a Partially Observable Markov Decision Process (POMDP) and propose the CGDM algorithm to efficiently generate optimal AI agent migration decisions.   
    \item In the CGDM algorithm, we introduce a denoising consistency term in the actor objective function to constrain the smoothness of the generated action distribution. we use it as a confidence measure to adaptively balance Q-value feedback with denoising loss during policy updates. Numerical results show that the CGDM algorithm significantly outperforms baseline methods in terms of both stability and performance.
\end{itemize}

The rest of the paper is organized as follows. Section~\ref{s2} summarizes the related work. Section~\ref{s3} introduces the proposed framework and formulates it as an optimization problem. In Section~\ref{s4}, we model the migration process as a POMDP and introduce the CGDM algorithm. Section~\ref{s5} presents numerical results, and Section~\ref{s6} concludes this paper.

\section{Related Work}\label{s2}
\subsection{Vehicular Metaverses}\label{s2-1}
The metaverse is an immersive ecosystem that integrates 3D virtual spaces and the real world~\cite{sami2024metaverse}, providing new application scenarios and exploration directions for various fields, such as ITSs, emerging the concept of vehicular metaverses~\cite{dwivedi2022metaverse}. The vehicular metaverse is not limited to a platform for virtual entertainment. Leveraging AI agents empowered by LLMs and VLMs, the vehicular metaverse is expected to advance ITSs toward a new paradigm of autonomous decision-making and real-time interaction. These AI agents are defined as digital entities that integrate perception, brain, and action modules~\cite{xi2025rise}, enabling them to perceive their environment, make decisions, and execute actions to achieve predefined goals. In~\cite{ma2024mobility}, vehicles are mapped to AI agents, synchronizing their physical and digital counterparts in real time. Consequently, vehicles evolve from mere transportation tools into intelligent decision-making entities capable of self-learning and interaction with environments. However, deploying AI agents to the vehicular metaverse also faces challenges in real-time data processing, dynamic environment adaptation, and privacy security. In~\cite{soliman2024artificial}, a multi-modal semantic perception framework was introduced that significantly reduced data transmission and latency in vehicular networks by extracting skeletal features and semantic information from images. Furthermore, in~\cite{wen2024defending}, the focus was on the security of AI agent migration, where the authors modeled network attacks against RSUs and proposed an online AI agent migration framework based on DRL and trust evaluation mechanisms, effectively defending against DDoS and malicious RSU attacks to achieve high-security AI agent migration in the vehicular metaverse.

\subsection{Service Migration}\label{s2-2}
Given that advanced AI systems (e.g., DTs or AI agents) are hosted on RSUs, when a vehicle leaves the coverage area of its current RSU, its AI systems need to be migrated to the next adjacent RSU to maintain continuous service and low latency. Existing works optimize service migration by proposing various evaluation metrics and methods, such as migration task freshness and latency~\cite{wen2023task,kang2024uav}, and balance migration latency with resource overhead by adopting methods such as contract models, dynamic resource scheduling, game theory, and DRL~\cite{zhong2025generative,chu2023metaslicing,kang2024metaverses}. In~\cite{chen2023multiagent}, the authors proposed a pre-migration strategy in which, before the vehicle leaves the coverage area of its current RSU, some or all of its service instances are replicated to a neighboring RSU. By predicting vehicle trajectories using long short-term memory and optimizing the pre-migration strategy with multi-agent DRL, the service migration latency was effectively reduced. However, existing works focus on making decisions from the perspective of vehicles or RSUs, ignoring the importance of bilateral collaboration in service migration.

Furthermore, frequent service migration in the vehicular metaverse brings potential security threats. In~\cite{luo2023privacy}, the author proposed a defense scheme based on dual pseudonyms and synchronous replacement for privacy attacks during the DT migration process, and combined blockchain and inventory theory to optimize pseudonym management, which significantly improved the effectiveness of privacy protection. To ensure large-scale and highly reliable service migration in the vehicular metaverse, the authors in~\cite{zhong2023blockchain} proposed a trust evaluation mechanism based on the subjective logic model, which effectively identified malicious RSUs. Although existing works have quantified the trustworthiness of RSUs from multiple perspectives, they overlook the individual differences in trust preference toward RSUs among vehicular users.

\subsection{Diffusion Models for Deep Reinforcement Learning}\label{s2-3}
Recently, diffusion models have received significant attention in the field of DRL. In the domain of offline RL, a policy optimization method based on the diffusion model has shown obvious advantages. The diffusion-based Q-Learning (QL) algorithm proposed by the authors in~\cite{DBLP:conf/iclr/WangHZ23} effectively alleviates the issue of policy distribution drift from the training data by integrating diffusion models with QL, achieving leading performance on the D4RL benchmark tasks. The diffusion Actor-Critic (AC) algorithm proposed in~\cite{fang2024diffusion} further optimizes the integration between diffusion models and the AC architecture, making the policy optimization more efficient and stable. 

In the domain of online RL, the application of diffusion models has also gradually attracted attention. The DIffusion POlicy (DIPO) algorithm proposed by the authors in~\cite{yang2023policy} first employed diffusion models for online policy optimization. By integrating action gradient updates with the enhanced representation capabilities of diffusion models for action distributions, DIPO outperforms traditional Gaussian-based policy methods. Subsequently, the authors in~\cite{ding2024diffusionbased} proposed the Q-weighted variational policy optimization algorithm. By designing a diffusion loss function based on Q-value weighting, the diffusion model can effectively integrate value information during online policy optimization and achieve better decision-making performance. In~\cite{wang2024diffusion}, the authors integrated diffusion policies with the AC architecture and addressed the difficulty of computing policy entropy through an entropy-regulation mechanism. However, the existing methods that combine diffusion models with online RL generally face the problem of unstable policy updates, which still needs to be solved.

\begin{figure*}[!htbp]
\centering
\includegraphics[width=0.85\textwidth]{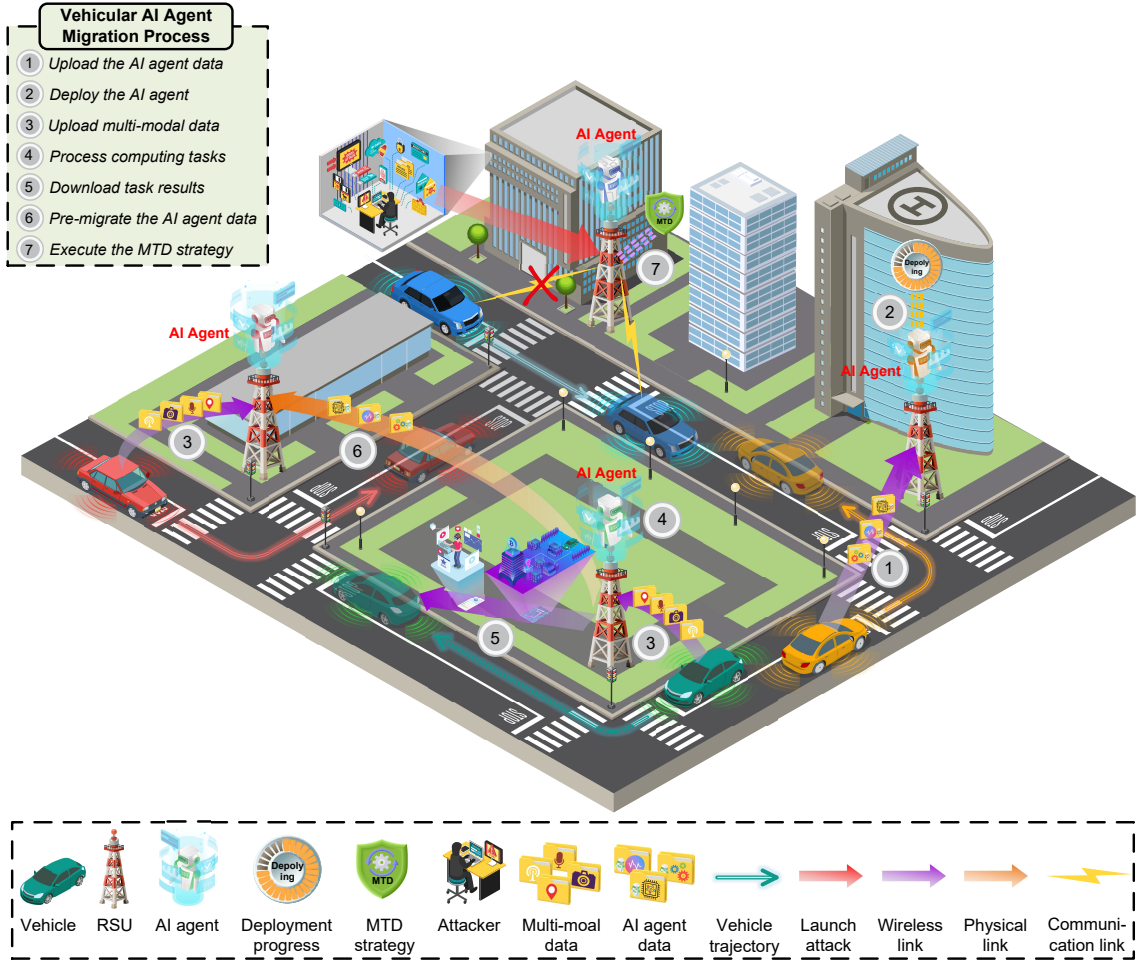}
\caption{A reliable vehicular AI agent migration framework in vehicular metaverses. When a vehicle enters the coverage area of an RSU that does not hold a replica of its AI agent, the vehicle uploads the data required to construct the AI agent. Then, the RSU deploys the AI agent and allocates the computing resources. While driving, the vehicle continuously uploads multi-modal data. After performing environment perception, decision-making, and action execution, the AI agent returns the resulting metaverse services to the vehicle. To ensure an immersive experience for users as vehicles move in real time, replicas of the AI agent are proactively pre-migrated to the next appropriate RSU. Moreover, the RSU can actively implement MTD strategies that dynamically reconfigure network addresses and topology, resulting in a brief link reestablishment.}
\label{fig1}
\end{figure*}

\section{Reliable Vehicular AI Agent Migration Framework}
\label{s3}
In this section, we first introduce how to migrate AI agents in the vehicular metaverse. Then, we propose a TPB-based trust evaluation model to address security threats during the migration process. To ensure low-latency and highly-reliable AI agent migration, we formulate this task as an optimization problem. The key notations in the paper are listed in Table~\ref{tab1}.

\subsection{Vehicular AI Agent Migration Model}
\label{s3-1}
We consider a vehicular metaverse consisting of multiple vehicles and multiple RSUs. As shown in Fig.~\ref{fig1}, each vehicle deploys its AI agent to an RSU and continuously uploads environmental data and user requests to the AI agent during driving. The AI agent processes these multi-modal data in real time to generate personalized vehicular metaverse services, which are then downloaded and executed by the vehicle. To ensure an immersive experience, the data required to construct the AI agent is pre-migrated to the next target RSU (i.e., the pre-migration process). The vehicle set is denoted as $\mathcal{V}=\{1,\ldots,v,\ldots,V\}$, the RSU set is denoted as $\mathfrak{Z}=\{1,\ldots,s,\ldots,S\}$, and time is divided into discrete time slots $\mathcal{T}=\{1,\ldots,t,\ldots,T\}$~\cite{he2022collaborative}.

Initially, we analyze the latency incurred when an AI agent is deployed on an RSU. Due to the continuous movement of vehicles, the Euclidean distance between the vehicle and the RSU changes dynamically. Specifically, the position of vehicle $v$ at time slot $t$ is denoted as $(x_v(t), y_v(t))$, and the fixed location of RSU $s$ is denoted as $(x_s, y_s)$. Thus, the Euclidean distance between vehicle $v$ and RSU $s$ can be calculated as $P_{v,s}(t)=\sqrt{[x_v(t)-x_s]^2+[y_v(t)-y_s]^2}$. Furthermore, the wireless communication channel between vehicles and RSUs is dynamically changing with distance. Considering uniform characteristics for both uplink and downlink wireless transmission channels~\cite{kang2024uav}, the path-loss channel gain between vehicle $v$ and RSU $s$ can be calculated as
\begin{equation}
    g_{v,s}(t)=\Gamma\left[\frac{c}{4\pi fP_{v,s}(t)}\right]^2,
\label{eq1}
\end{equation}
where $\Gamma$ denotes the channel gain coefficient, $c$ is the speed of light, and $f$ is the carrier frequency. When a vehicle enters the coverage area of an RSU, if it lacks a copy of the corresponding AI agent, the vehicle needs to upload the data required to construct its AI agent (e.g., model parameters and status information). To mitigate interference from other vehicles transmitting data, we employ orthogonal frequency division multiple access to allocate orthogonal subcarriers to different vehicles~\cite{tang2020deep}. According to the Shannon-Hartley formula~\cite{shannon1949communication}, the uplink transmission rate from vehicle $v$ to RSU $s$ can be calculated as
\begin{equation}
    R_{v,s}^{up}(t)=B^{up}\log_2\left[1+\frac{h_vg_{v,s}(t)}{N_s^2}\right],
\label{eq2}
\end{equation}
where $B^{up}$ is the uplink bandwidth, $h_v$ is the transmission power of vehicle $v$, and $N_s^2$ is the additive Gaussian white noise at RSU $s$. Thus, the latency for uploading the data $D^{con}_{v}$ required to construct the AI agent can be calculated as
\begin{equation}
    T_{v,s}^{uc}(t)=\frac{D^{con}_{v}(t)}{R_{v,s}^{up}(t)}.
\label{eq3}
\end{equation}

Upon receiving the data required to construct the AI agent, the RSU $s$ performs the deployment operations. These include loading the corresponding model, synchronizing the state, and allocating computing resources to ensure that the AI agent can promptly respond when a user initiates a service request. The latency caused by deploying the AI agent for vehicle $v$ can be calculated as
\begin{equation}
    T_{v,s}^{dep}(t)=\frac{d_sD^{con}_v(t)}{\alpha_{v,s}(t)C_s},
\label{eq4}
\end{equation}
where $\alpha_{v,s}(t) \in (0,1)$ is the ratio of computing resources allocated by RSU $s$ to the AI agent corresponding to vehicle $v$, $d_s$ and $C_s$ denote the number of CPU cycles required by RSU $s$ to process one unit of data and its CPU speed, respectively. To realize immersive vehicular metaverse services, vehicle $v$ continuously collects environmental information (e.g., real-time images and radar data) and user interaction requirements (e.g., AR navigation requests) during driving, and uploads these multi-modal data to the AI agent deployed on RSU $s$. The resulting uplink latency can be calculated as
\begin{equation}
    T_{v,s}^{um}(t)=\frac{D_{v}^{raw}(t)}{R_{v,s}^{up}(t)},
\label{eq5}
\end{equation}
where $D_{v}^{raw}(t)$ represents the size of raw multi-modal data collected by the vehicle $v$ at time slot $t$. After the AI agent receives the uploaded multi-modal data, it first performs data preprocessing, feature extraction, and environment perception through the perception module. Then the brain module generates decisions based on these processed data, e.g., optimizing the navigation route or producing virtual interactive content. Subsequently, the action module executes tasks according to these decisions, e.g., computing the optimal driving route or rendering AR virtual scenes. The computing load incurred by the AI agent during environment perception, decision-making, and action execution is represented as $D^{com}_v$, and its processing latency can be calculated as
\begin{equation}
    T_{v,s}^{pro}(t)=\frac{d_sD_v^{com}(t)}{\alpha_{v,s}(t)C_s}.
\label{eq6}
\end{equation}

After the AI agent completes its computing task, the generated results, such as navigation instructions or rendered frames, are transmitted back to the vehicle through the wireless link between the RSU and the vehicle, thereby enabling users to experience the metaverse services through XR devices. Similar to the uplink rate, the downlink rate between the vehicle $v$ and the RSU $s$ is calculated as $R_{v,s}^{down}(t)=B^{down}\mathrm{log}_2[1+\frac{p_vg_{v,s}(t)}{N_s^2}]$. Thus, the latency of the vehicle $v$ downloading the result data $D_v^{res}$ can be calculated as
\begin{equation}
    T_{v,s}^{down}(t)=\frac{D_v^{res}(t)}{R_{v,s}^{down}(t)}.
\label{eq7}
\end{equation}

Due to the limited coverage area of a single RSU, before vehicle $v$ leaves the coverage area of the current RSU $s$, the data required to construct the AI agent should be pre-migrated to the next appropriate RSU $s_p$, thus avoiding the long latency caused by repeatedly uploading large volumes of data to recreate a AI agent replica. When the vehicle enters the coverage of RSU $s_p$, RSU $s_p$ can directly deploy the AI agent for vehicle $v$, ensuring service continuity. The latency caused by the pre-migration process can be calculated as
\begin{equation}
    T_{s,s_p}^{mig}\left(t\right)=\frac{D^{con}_v(t)}{B_{s,s_p}},
\label{eq8}
\end{equation}
where $B_{s,s_p}$ is the physical link bandwidth between RSU $s$ and RSU $s_p$. Considering potential security threats to RSUs, such as attackers launching DDoS attacks to interrupt data exchanges between RSUs and vehicles~\cite{he2021game}, an RSU may opt to implement MTD strategies~\cite{zhang2022mitigate}. These strategies include dynamically updating the network topology, reconfiguring network addresses, or executing secure handshakes and key updates, all designed to actively disrupt the ability of attackers to identify and track the network environment, thereby reducing the security risk. However, when implementing the MTD strategy, RSU $s$ need to redeploy its internal configurations and reestablish communication links~\cite{zhang2023moving}, which causes a brief latency in the communication between vehicles and RSUs, consequently introducing an additional latency $T_{v,s}^{ext}$ in the vehicular AI agent migration.

\begin{table}[!t]
\centering
\caption{Key Notations in the Paper}
\label{tab1}
\begin{tabular}{>{\centering\arraybackslash}p{2.1cm}>{\raggedright\arraybackslash}p{5.4cm}} \toprule
 \textbf{Parameters}& \textbf{Descriptions} \\ \midrule
$\mathcal{A}_{u,s}$&The attitude factor of user $u$ to RSU $s$\\
$\mathcal{B}_{u,s}$&Reputation value of RSU $s$ for user $u$\\
$B_{s,s_p}$& Physical link bandwidth between RSU $s$ and RSU $s_p$\\
$C_s$&CPU speed of RSU $s$\\
$\mathcal{M}_s$&Migration efficiency of RSU $s$\\
$\mathcal{P}_{u,s}$&The perceived behavioral control factor of user $u$ to RSU $s$\\
$\mathcal{R}_s$&Data transmission reliability of RSU $s$\\
$R^{up}_{v,s},R^{down}_{v,s}$&Uplink and downlink transmission rates between vehicle $v$ and RSU $s$, respectively\\
$\mathcal{S}_{u,s}$&The subjective norm factor of user $u$ to RSU $s$\\
$T^{uc}_{v,s}, T^{dep}_{v,s}$&Latency caused by uploading and deploying AI agent, respectively\\
$T^{um}_{v,s}$&Latency caused by uploading multi-modal data\\
$T^{pro}_{v,s}$&Latency caused by processing computing tasks\\
$T^{down}_{v,s}$&Latency caused by downloading task results\\
$T^{mig}_{v,s}$&Latency caused by pre-migrating AI agent\\
$T^{ext}_{v,s}$&Latency caused by RSU $s$ executing MTD strategies\\
$T^{tot}_{v}$&Total latency for the vehicular AI agent migration\\
$\mathcal{U},\mathcal{E}$&Set of vehicular metaverse users and their interaction evaluations, respectively\\
$\mathcal{V}, \mathcal{S},\mathcal{T}$&Set of vehicles, RSUs and time slots, respectively\\
$\vartheta_{u,s}$&The probability that user $u$ gives a positive evaluation to RSU $s$\\
\bottomrule
\end{tabular}
\end{table}

In summary, to provide users with immersive metaverse services, vehicle $v$ first uploads the data required to construct its AI agent to RSU $s$ and completes the deployment, which incurs uplink latency $T_{v,s}^{uc}(t)$ and deployment latency $T_{v,s}^{dep}(t)$. Then, vehicle $v$ continuously transmits the collected multi-modal data, and the AI agent performs data preprocessing, feature extraction, and decision generation, as well as executing corresponding tasks, resulting in uplink latency $T_{v,s}^{um}(t)$ and processing latency $T_{v,s}^{pro}(t)$. Simultaneously, to ensure service continuity, the AI agent proactively pre-migrates a replica of itself to the next target RSU $s_p$, resulting in migration latency $T_{s,s_p}^{mig}$. After the computing task is completed, the results generated by the AI agent are transmitted back to vehicle $v$ through the downlink, resulting in a downlink latency $T_{v,s}^{down}(t)$. Additionally, RSU $s$ can choose to execute MTD strategy to enhance security, which introduces an extra latency $T_{v,s}^{down}(t)$. Therefore, the total latency for the vehicular AI agent migration can be calculated as
\begin{equation}
\begin{aligned}
    T_{v}^{tot}(t)&=(1-\varepsilon_s)(T_{v,s}^{uc}(t)+T_{v,s}^{dep}(t))+T_{v,s}^{um}(t)+T_{v,s}^{down}(t) \\
    &\quad+\max\left\{T_{v,s}^{pro}(t),T_{s,s_{p}}^{mig}(t)\right\}+\beta_{s}(t)T_{v,s}^{ext}(t),
\end{aligned}
\label{eq9}
\end{equation}
where $\varepsilon_s \in \{0,1\}$ indicates whether RSU $s$ has a copy of the AI agent corresponding to vehicle $v$. Specifically, $\varepsilon_s=1$ if the RSU $s$ has the AI agent copy, and $\varepsilon_s=0$ otherwise. Additionally, $\beta_{s}(t)\in \{0,1\}$ indicates whether RSU $s$ executes the MTD strategy at time slot $t$, where $\beta_{s}(t)=1$ indicates execution of the MTD strategy, and $\beta_{s}(t)=0$ otherwise.

\subsection{TPB-based Trust Evaluation Model}
\label{s3-2}
To address the security threats caused by cyber attacks during vehicular AI agent migration, RSUs can proactively implement MTD strategies to reduce the possibility of cyber attacks. However, relying solely on proactive defenses at the RSU side is insufficient to eliminate all potential threats. Once an RSU fails to execute the MTD strategy in time and becomes compromised, vehicles that continue to migrate their AI agents to that RSU will face significant security risks. Therefore, it is necessary for vehicles to conduct trust evaluations of RSUs to identify those that have been compromised or have potential risks in real time.

Considering that existing trust evaluation models fail to account for differences in user preferences regarding security risks and service quality~\cite{john2024trust}, and lack a comprehensive characterization of multi-source trust variability, we propose a TPB-based trust evaluation model. This model not only objectively quantifies RSU security but also considers the differences in trust preferences among vehicle occupants (i.e., vehicular metaverse users), thereby providing highly reliable and personalized trust evaluation results for vehicular AI agent migration decisions.

According to the TPB~\cite{AJZEN1991179}, the intention of an individual to perform a certain behavior is mainly determined by the attitude $\mathcal{A}$, the subjective norm $\mathcal{S}$, and the perceived behavioral control $\mathcal{P}$. Specifically, $\mathcal{A}$ reflects the positive or negative evaluation of the behavior itself, $\mathcal{S}$ reflects the social pressure and support perceived during the decision-making process, and $\mathcal{P}$ reflects the perception of the ease or difficulty of performing the behavior. We define the vehicular metaverse user set as $\mathcal{U}=\{1,\ldots,u,\ldots,U\}$, which corresponds one-to-one to the vehicle set $\mathcal{V}=\{1,\ldots,v,\ldots,V\}$. Moreover, we define the reputation value $\mathcal{B}_{u,s}$ of RSU $s$, indicating the behavioral intention of user $u$ to migrate its AI agent to RSU $s$.

\textit{1) Attitude $\mathcal{A}$}

The attitude factor is the subjective preference of users towards an RSU, measured by their historical interaction evaluation. Let $\mathcal{E}=\left\{e_{u,s}^{n}|u\in\mathcal{U},s\in\mathfrak{Z},n=1,\dots,n_{u,s}\right\}$ represent the interaction evaluation set between users and RSUs, where $n_{u,s}$ is the total number of interactions between user $u$ and RSU $s$, and $e_{u,s}^{n}\in\{0,1\}$ represents the $n$-th evaluation to RSU $s$ by user $u$. Here, $e_{u,s}^{n}=1$ indicates a positive evaluation, whereas $e_{u,s}^{n}=0$ indicates a negative evaluation. Since these evaluations are binary, they can be modeled using a Bernoulli distribution with parameters $\vartheta_{u,s}$, i.e., $ e_{u,s}^n\sim \mathrm{Ber}(\vartheta_{u,s})$, where $\vartheta_{u,s}$ denotes the probability that user $u$ gives a positive evaluation. To estimate $\vartheta_{u,s}$ with limited evaluation data, let the Beta distribution as the prior~\cite{el2020design}, which can be expressed as
\begin{equation}
\vartheta_{u,s}\thicksim \mathrm{Beta}(\alpha_{\mathcal{A}},\beta_{\mathcal{A}}),
\label{eq10}
\end{equation}
where $\alpha_{\mathcal{A}},\beta_{\mathcal{A}}>0$ are adjustable prior hyperparameters of $\vartheta_{u,s}$. Owing to the conjugate relationship between the Bernoulli and Beta distributions, after obtaining the interactive evaluation set $\mathcal{E}_{u,s}$ between user $u$ and RSU $s$, the posterior distribution of $\vartheta_{u,s}$ remains a Beta distribution, which can be expressed as
\begin{equation}
\vartheta_{u,s}|\mathcal{E}_{u,s}\thicksim \mathrm{Beta}(\alpha_{\mathcal{A}}+p_{u,s},\beta_{\mathcal{A}}+q_{u,s}),
\label{eq11}
\end{equation}
where $p_{u,s}=\sum_{n=1}^{n_{u,s}}e_{u,s}^{n}$ and $q_{u,s}=n_{u,s}-p_{u,s}$ are the total numbers of positive and negative evaluations by user $u$ for RSU $s$, respectively. Taking the expectation of the posterior distribution of $\vartheta_{u,s}$ as the subjective favorability of user $u$ to RSU $s$, the attitude factor can be calculated as
\begin{equation}
\mathcal{A}_{u,s}=\mathbb{E}[\vartheta_{u,s}|\mathcal{E}_{u,s}]=\frac{\alpha_{\mathcal{A}}+p_{u,s}}{\alpha_{\mathcal{A}}+\beta_{\mathcal{A}}+p_{u,s}+q_{u,s}}.
\label{eq12}
\end{equation}

\textit{2) Subjective Norm $\mathcal{S}$}

The subjective norm factor describes the influence of the external environment on user behavior, reflecting the guiding role of group consensus on user trust judgment. It is measured by the overall evaluation of other users on RSU $s$. Similar to the quantification of attitude factor, we also use a Beta distribution as the prior distribution for the probability $\vartheta_{\neg u,s}$ that other users give positive evaluations, i.e., $\vartheta_{\neg u,s}\thicksim \mathrm{Beta}(\alpha_{\mathcal{S}},\beta_{\mathcal{S}})$, where $\alpha_{\mathcal{S}},\beta_{\mathcal{S}}>0$ are prior hyperparameters of $\vartheta_{\neg u,s}$. After obtaining the interaction evaluation set $\mathcal{E}_{\neg u,s}=\{e_{u^{\prime},s}^n|u^{\prime}\in(\mathcal{U}\backslash\{u\}),n=1,\dots,n_{u^{\prime},s}\}$ by other users for RSU $s$, and leveraging the conjugacy between Beta and Bernoulli distribution, the posterior distribution of $\vartheta_{\neg u,s}$ is given by
\begin{equation}
\vartheta_{\neg u,s}|\mathcal{E}_{\neg u,s}\thicksim \mathrm{Beta}(\alpha_{\mathcal{S}}+p_{\neg u,s},\beta_{\mathcal{S}}+q_{\neg u,s}),
\label{eq13}
\end{equation}
where $p_{\neg u,s}=\sum_{u^{\prime}\in(\mathcal{U}\setminus\{u\})}\sum_{n=1}^{n_{u^{\prime},s}}e_{u^{\prime},s}^{n}$ and $q_{\neg u,s}=\sum_{u^{\prime}\in(\mathcal{U}\setminus\{u\})}n_{u^{\prime},s}-p_{\neg u,s}$ are the total number of positive and negative evaluations on RSU $s$ by other users, respectively. 

Therefore, the subjective norm factor is quantified as the expectation of the posterior distribution of $\vartheta_{\neg u,s}$. By updating $p_{\neg u,s}$ and $q_{\neg u,s}$ to reflect the trust changes of other users on RSU $s$, the subjective norm factor can be calculated as
\begin{equation}
\mathcal{S}_{u,s}=\mathbb{E}[\vartheta_{\neg u,s}|\mathcal{E}_{\neg u,s}]=\frac{\alpha_{\mathcal{S}}+p_{\neg u,s}}{\alpha_{\mathcal{S}}+\beta_{\mathcal{S}}+p_{\neg u,s}+q_{\neg u,s}}.
\label{eq14}
\end{equation}

\textit{3) Perceived Behavioral Control $\mathcal{P}$}

The perceived behavioral control factor describes the perception of users regarding whether RSUs can complete the AI agent migration tasks smoothly and quickly. This factor is quantified by evaluating the reliability and efficiency of RSUs during vehicular AI agent migration. To evaluate the data transmission reliability of RSU $s$, we use the packet statistics recorded in the beacon messages periodically broadcast by RSU $s$ and vehicles during data transmission. Specifically, let $D_s$and $D_f$ represent the numbers of packets successfully and unsuccessfully received by RSU $s$, reflecting its ability to obtain multi-modal raw data from vehicles. Let $F_s$ and $F_f$ represent the numbers of packets successfully and unsuccessfully forwarded by RSU $s$, indicating its capacity to deliver processed results to vehicles. Based on the above beacon messages, the packet delivery rate $\mathcal{R}_{s}^{D}$ and the forwarding rate $\mathcal{R}_{s}^{F}$ of RSU $s$ can be calculated as
\begin{equation}
\mathcal{R}_{s}^{D}=\frac{D_{s}-D_{f}}{D_{s}+D_{f}},~\mathcal{R}_{s}^{F}=\frac{F_{s}-F_{f}}{F_{s}+F_{f}}.
\label{eq15}
\end{equation}

Since the data exchange between the RSU and vehicles is bidirectional, both the uplink and downlink transmission processes must be considered. Therefore, the overall data transmission reliability of RSU $s$ is defined as
\begin{equation}
\mathcal{R}_s=[\lambda\mathcal{R}_s^D+(1-\lambda)\mathcal{R}_s^F],
\label{eq16}
\end{equation}
where $\lambda \in(0,1)$ is the weight coefficient that adjusts the relative importance of $\mathcal{R}_s^D$ and $\mathcal{R}_s^F$. Next, to evaluate the efficiency of RSU $s$ during the vehicular AI agent migration, we consider the set of total migration latency $\mathcal{L}_s=\{T_{j}^{tot}(t)|j\in \mathcal{V}_s(t), t\in(T^{\prime}-\tau,T^{\prime})\}$ associated with RSU $s$, where the set of vehicles whose AI agents are deployed on RSU $s$ at time slot $t$ is denoted by $\mathcal{V}_s(t)$, $T'$ is the current time, and $\tau$ is the length of the time window. The average total migration latency at RSU $s$ in this time window can be calculated as
\begin{equation}
\overline{T_s^{tot}}=\frac{1}{|\mathcal{L}_s|}\sum_{t\in(T^{\prime}-\tau,T^{\prime})}\sum_{j\in \mathcal{V}_s(t)}T_{j}^{tot}\left(t\right).
\label{eq17}
\end{equation}

To reflect the personalized requirements of different users for migration efficiency, we introduce the maximum latency that user $u$ can accept, i.e., tolerable latency $T_u^{max}$, and define the migration efficiency of RSU $s$ as
\begin{equation}
\mathcal{M}_s 
= \max\!\left\{0,\;1 - \frac{\overline{T_s^{\mathrm{tot}}}}{T_u^{\max}}\right\}.
\label{eq18}
\end{equation}

When the average total migration latency $\overline{T_s^{tot}}$ is significantly lower than the tolerable latency $T_u^{max}$, $\mathcal{M}_s$ approaches $1$, indicating that the migration efficiency of RSU $s$ fully satisfies the latency requirement of user $u$. As the average migration latency $\overline{T_s^{tot}}$ gradually approaches the tolerable latency ${T_u^{max}}$, the satisfaction of user $u$ with the migration efficiency of RSU $s$ decreases. Once the average migration latency $\overline{T_s^{tot}}$ exceeds the tolerable latency $T_u^{max}$, $\mathcal{M}_s$ drops to $0$, indicating that the migration efficiency of RSU $s$ is unacceptable for user $u$. Finally, by taking a weighted sum of data transmission reliability $\mathcal{R}_s$ and migration efficiency $\mathcal{M}_s$, the perceived behavioral control factor is given by
\begin{equation}
\mathcal{P}_{u,s}=\sigma\mathcal{R}_s+(1-\sigma)\mathcal{M}_s,
\label{eq19}
\end{equation}
where $\sigma \in(0,1)$ is used to adjust the relative importance between $\mathcal{R}_s$ and $\mathcal{M}_s$.

\textit{4) Behavioral Intention as Reputation Value}

After obtaining the attitude factor $\mathcal{A}_{u,s}$, the subjective norm factor $\mathcal{S}_{u,s}$, and the perceived behavior control factor $\mathcal{P}_{u,s}$, the behavioral intention of user $u$ to migrate the AI agent to RSU $s$ can be quantified according to TPB, i.e., the reputation value of RSU $s$ for user $u$, which is calculated as
\begin{equation}
\mathcal{B}_{u,s}=\zeta_{\mathcal{A}}\mathcal{A}_{u,s}+\zeta_{\mathcal{S}}\mathcal{S}_{u,s}+\zeta_{\mathcal{P}}\mathcal{P}_{u,s},
\label{eq20}
\end{equation}
where $\zeta_{\mathcal{A}},\zeta_{\mathcal{S}},\zeta_{\mathcal{P}}\in(0,1)$ satisfy $\zeta_{\mathcal{A}}+\zeta_{\mathcal{S}}+\zeta_{\mathcal{P}}=1$. These weights can be allocated based on the relative importance of $\mathcal{A}_{u,s}$, $\mathcal{S}_{u,s}$, and $\mathcal{P}_{u,s}$.

Since the state of RSU $s$ is dynamically changing, a single trust evaluation is often insufficient to reflect its long-term stability. Therefore, the reputation value $\mathcal{B}_{u,s}$ need to be updated smoothly to integrate both historical evaluations and the latest data.  Let $\mathcal{B}_{u,s}^{near}$ denote the instantaneous reputation value obtained by evaluating the latest data, and let $\mathcal{B}_{u,s}^{old}$ denote the reputation value from the previous time slot. Then, the updated reputation value at the current moment can be calculated as
\begin{equation}
\mathcal{B}_{u,s}^{new}=\xi \mathcal{B}_{u,s}^{near}+(1-\xi)\mathcal{B}_{u,s}^{old},
\label{eq21}
\end{equation}
where $\xi \in(0,1)$ controls the update rate.

\subsection{Problem Formulation}
\label{s3-3}

To achieve low latency and high reliability of vehicular AI agent migration while optimizing the decision variable $\mathbb{A}$ under the constraints of limited RSU resources and security requirements, the optimization problem is formulated as
\begin{subequations}
\begin{align}
\min_{\mathbb{A}} & \sum_{t=1}^{T} \sum_{v=1}^{V} T^{tot}_{v}(t) \label{eq22a}\\
\rm{s.t.} \quad & \sum_{j=1}^{|\mathcal{V}_s(t)|}\alpha_{j,s}(t)\leq1,\forall s\in\mathfrak{Z}, \forall t \in \mathcal{T}, \label{eq22b}\\
& L_s(t)\leq L_s^{max},\forall s\in \mathfrak{Z}, \forall t \in \mathcal{T}, \label{eq22c}\\
& \mathcal{B}_{v,s_p}(t)\geq\mathcal{B}^{thre},\forall v\in V, \forall t \in \mathcal{T}. \label{eq22d}
\end{align}
\end{subequations}

Constraint~\eqref{eq22b} limits the computing resources allocated by each RSU to not exceed its total computing resources. Constraint~\eqref{eq22c} ensures that the load $L_s(t)$ of each RSU is below its maximum load capacity $L_s^{max}$. Finally, constraint~\eqref{eq22d} requires that the reputation value of the target RSU $s_p$ selected for the AI agent in the pre-migration process should exceed a safety threshold $\mathcal{B}^{thre}$, thereby ensuring reliable vehicular AI agent migration.

\section{Confidence‑Regulated Generative Diffusion Models for Vehicular AI Agent Migration}
\label{s4}

In this section, we model vehicular AI agent migration as a POMDP and propose a CGDM algorithm that combines DRL and diffusion models to generate optimal decisions.

\subsection{POMDP Modeling}
\label{s4-1}

For vehicular AI agent migration, the objective is to minimize the total latency in~\eqref{eq22a}. Given that the network environment is highly dynamic~\cite{du2024diffusion}, with migration decisions intricately coupled to RSU load conditions and potential security risks arising from cyber attacks~\cite{zhang2022mitigate}, we model the vehicular AI agent migration process as a POMDP as follows:

\textit{1) Observation Space}

The observation space $\mathbb{O}$ consists of the environmental information that vehicles and RSUs can obtain in each time slot, which is defined as $\mathbb{O}(t)=\{P_{\mathcal{V}}(t),K_{\mathcal{V}}(t),L_{\mathfrak{Z}}(t),\mathcal{B}(t)\}$. Here, $P_{\mathcal{V}}(t)=\{(x_v,y_v)| v \in \mathcal{V}\}$ is the position set of all vehicles, reflecting the communication distances and link qualities between vehicles and RSUs. $K_{\mathcal{V}}(t)=\{K_v(t)\mid v\in\mathcal{V},K_v(t)\in\{1,2,\ldots,S\}\}$ is the index set of the RSUs on which AI agents are deployed. These indexes are determined by the pre-migration decision of vehicles in the previous time slot and are used to help RSUs to allocate computing resources. $L_{\mathfrak{Z}}(t)=\{L_s(t)|s\in\mathfrak{Z}\}$ is the load set of all RSUs, which reflects the current computing burden on each RSU. $\mathcal{B}(t)=\{\mathcal{B}_{u,s}(t)|u\in\mathcal{U},s\in\mathfrak{Z}\}$ is the reputation value set of each RSU for different users, which helps vehicles in pre-migrating their AI agents to reliable RSUs.

\textit{2) Action Space}

The action space is defined as the decision set that the DRL agent makes for all vehicles and RSUs at each time slot, denoted as $\mathbb{A}=\{A_{\mathcal{V}},A_{\mathfrak{Z}}\}$. According to Section~\ref{s3}, the decisions regarding the selection of a pre-migration RSU for vehicles and the decision on whether an RSU executes the MTD strategy are discrete, while the decision for RSU resource allocation is continuous, resulting in a hybrid action space. To avoid the complexities and instability in policy gradient calculation caused by the hybrid action space~\cite{eisenach2018marginal}, we express all decisions as continuous actions and map them to discrete or continuous decisions when executing the decisions.

Specifically, the DRL agent makes a pre-migration decision $A_{\mathcal{V}}(t)=\{a_v(t)|v\in\mathcal{V}\}$ for all vehicles at time slot $t$, where the action vector $a_v(t)=[a_{v,1}(t),a_{v,2}(t),\ldots,a_{v,S}(t)]\in[0,1]^S$ of vehicle $v$ is converted into a probability distribution of each RSU as a pre-migration target through a softmax operation, and the RSU with the highest probability is selected as the migration target when executing the decision. Concurrently, the DRL agent makes resource allocation and defense strategies $A_{\mathfrak{Z}}(t)=\{(\alpha_s(t),\beta_s(t))|s\in\mathfrak{Z}\}$ for all RSUs, where $\alpha_s(t)=\{\alpha_{j,s}(t)\mid j\in\mathcal{V}_s(t),\alpha_{j,s}(t)\in(0,1)\}$ is the computing resource allocation ratio of RSU $s$ to the vehicle set $\mathcal{V}_s(t)$ it is currently serving, satisfying the total resource constraint~\eqref{eq22b}. Moreover, $\beta_s(t)\in[0,1]$ indicates the probability that RSU $s$ executes the MTD strategy. When executing the decision, $\beta_s(t)$ is a threshold such that if $\beta_s(t)>0.5$, it is mapped to $1$, indicating the execution of the MTD strategy, otherwise, $\beta_s(t)$ is mapped to $0$.

\textit{3) Transition Function}

The transition function characterizes the probability distribution of transitioning from state $\mathbb{O}(t)$ to $\mathbb{O}(t+1)$ after taking action $\mathbb{A}(t)$. Due to the highly dynamic and uncertain environment, it is difficult to formulate this function mathematically. Specifically, the positions of vehicles are determined by their own motion pattern and change continuously in each time slot, thus influencing the communication link quality between vehicles and RSUs. The load state of RSUs is affected by the data volume uploaded by vehicles, resource allocation strategies, and network transmission conditions. Concurrently, the reputation values of RSUs are dynamically updated based on their performance in data transmission and task processing, as well as user feedback. Additionally, potential cyber attacks may impact the migration process of AI agents, further increasing the complexity of state transitions. All these factors collectively determine how the state transitions over time.

\textit{4) Reward Function}

The reward represents the immediate return that the DRL agent obtains by executing action $\mathbb{A}(t)$ in state $\mathbb{O}(t)$. To minimize the total latency of vehicular AI agent migration, the immediate reward at time slot $t$ is defined as
\begin{equation}
\mathbb{R}\left(\mathbb{O}\left(t\right),\mathbb{A}\left(t\right)\right)=-\sum_{v=1}^VT_v^{tot}(t).
\label{eq23}
\end{equation}

\subsection{Confidence‑Regulated Generative Diffusion Models for Generating Optimal Decisions}
\label{s4-2}

AI agent migration problems typically exhibit highly dynamic and nonlinear characteristics, resulting in a complex, multi-modal decision distribution~\cite{du2024enhancing}. However, most DRL algorithms rely on unimodal distributions (e.g., Gaussian distributions) for policy representation~\cite{huang2023reparameterized}, which inadequately capture the subtle differences among multiple potential optimal decisions in the environment and are prone to converge to sub-optimal solutions. In contrast, diffusion policies for DRL can capture the multi-modal structure of the decision distribution through a denoising process~\cite{DBLP:conf/iclr/WangHZ23}, effectively representing the diversity and complexity of policy distributions. Therefore, we propose the CGDM algorithm for generating optimal AI agent migration decisions.

\begin{figure*}[!ht]
\centering
\includegraphics[width=0.85\textwidth]{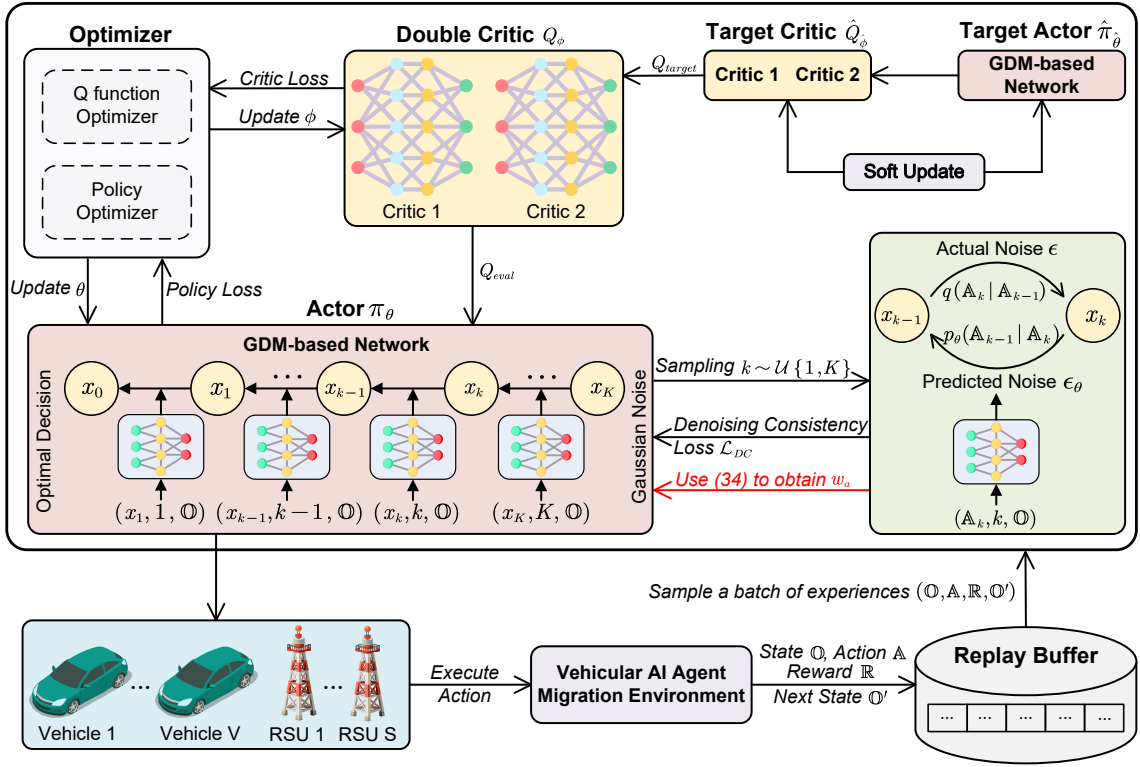}
\caption{The overall architecture of the CGDM algorithm. The actor network $\pi_{\theta}$ generates action by performing $K$ denoising steps on Gaussian noise. The resulting state–action pairs are then evaluated by the double critic networks $Q_\phi$ to guide policy optimization. A denoising consistency loss is computed to measure the confidence to adaptively adjust the optimization objective of the actor. All interaction samples are stored in the experience replay buffer for updates of both the actor and critic networks. To ensure stable training, the target networks are softly updated at each iteration.}
\label{fig3}
\end{figure*}

In the following, we first introduce the forward and reverse processes of the diffusion model and then describe the algorithmic architecture of CGDM.

\textit{1) Forward Process}
\label{s4-2-1}

The forward process in the diffusion model refers to the process of gradually adding Gaussian noise to the original data, which is usually modeled as a Markov chain of length $T$. At each time step $k=1,2,\dots,K$, Gaussian noise is added to the data distribution $x_{k-1}$ from the previous time step to obtain the data distribution $x_k$. The conditional probability distribution of a single-step transfer is defined as a normal distribution with a mean of $\sqrt{1-\beta_{k}}x_{k-1}$ and a variance of $\beta_k$, expressed as~\cite{ho2020denoising}
\begin{equation}
q(x_k\mid x_{k-1})=\mathcal{N}\left(x_k;\sqrt{1-\beta_k}x_{k-1},\beta_k\mathbf{I}\right),
\label{eq24}
\end{equation}
where $\beta_k$ controls the strength of the noise added at the time step $k$ and is determined by a variational scheduler. 

Since the forward process satisfies the Markov property, the state of each step depends only on the state of the previous time step. Therefore, the joint probability distribution for transitioning from the original data distribution $x_0$ to the final data distribution $x_K$ can be expressed as~\cite{ho2020denoising}
\begin{equation}
q(x_{1:K}\mid x_0)=\prod_{k=1}^Kq(x_k\mid x_{k-1}).
\label{eq25}
\end{equation}

Based on~\eqref{eq25}, we can further derive a direct expression for the data distribution at any time step $k$ relative to the original data distribution $x_0$. By defining $\alpha_k = 1-\beta_k$ and the cumulative product $\bar{\alpha}_{k}=\prod_{i=1}^{k}\alpha_{i}$, we obtain~\cite{ho2020denoising}
\begin{equation}
x_k=\sqrt{\bar{\alpha}_k}x_0+\sqrt{1-\bar{\alpha}_k}\epsilon,
\label{eq26}
\end{equation}
where $\epsilon\sim \mathcal{N}(0,\mathbf{I})$. When $K$ is large enough, $x_K$ almost loses the original data distribution structure and is close to an isotropic standard normal distribution $\mathcal{N}({0,\mathbf{I}})$. This property facilitates the derivation and formulation of the reverse process in the diffusion model.

\textit{2) Reverse Process}
\label{s4-2-2}

The objective of the reverse process is to gradually recover samples that approximate the original data distribution from pure noise. Specifically, under the condition of the environmental state $\mathbb{O}$, a noise $x_K \sim \mathcal{N}(0,\mathbf{I})$ is first sampled from the standard normal distribution. Subsequently, noise is gradually removed through the reverse process until the final decision $x_0$ is recovered. The reverse process can be modeled as a Markov chain comprising $K$ conditional probabilities, with joint distribution expressed as
\begin{equation}
p_\theta(x_{0:K})=p(x_K)\prod_{k=1}^Kp_\theta(x_{k-1}\mid x_k),
\label{eq27}
\end{equation}
where the reverse conditional probability $p_\theta(x_{k-1}\mid x_k)$ is not directly available but can be approximated by a parameterized model that learns its mean and variance, expressed as
\begin{equation}
p_\theta(x_{k-1}\mid x_k)=\mathcal{N}\left(x_{k-1};\mu_\theta(x_k,k,\mathbb{O}),\tilde{\beta}_k\mathbf{I}\right),
\label{eq28}
\end{equation}
where the variance $\tilde{\beta}_k=\frac{1-\bar{\alpha}_{k-1}}{1-\bar{\alpha}_k}\beta_k$ is set as a fixed time-dependent constant according to the Denoising Diffusion Probabilistic Model (DDPM)~\cite{ho2020denoising}. The mean $\mu_\theta(x_k,k,\mathbb{O})$ is learned through a neural network. By applying the Bayesian formula, the posterior distribution can be obtained as
\begin{equation}
q(x_{k-1}\mid x_k,x_0)=\mathcal{N}\left(x_{k-1};\tilde{\mu}(x_k,x_0,k),\tilde{\beta}_t\mathbf{I}\right),
\label{eq29}
\end{equation}
where the posterior mean $\tilde{\mu}(x_k,x_0,k)$ is computed by
\begin{equation}
\tilde{\mu}(x_k,x_0,k)=\frac{\sqrt{\bar{\alpha}_{k-1}}\beta_k}{1-\bar{\alpha}_k}x_0+\frac{\sqrt{\alpha_k}(1-\bar{\alpha}_{k-1})}{1-\bar{\alpha}_k}x_k.
\label{eq30}
\end{equation}

Since the decision $x_0$ is unknown during inference, we employ a denoising network $\epsilon_\theta(x_k,t,\mathbb{O})$ to predict the noise contained in $x_k$. By substituting~\eqref{eq26} into ~\eqref{eq30} to obtain an approximate posterior mean, given by
\begin{equation}
\mu_\theta(x_k,k,\mathbb{O})=\frac{1}{\sqrt{\alpha_k}}\left(x_k-\frac{\beta_k\tanh\left(\epsilon_\theta(x_k,k,\mathbb{O})\right)}{\sqrt{1-\bar{\alpha}_k}}\right),
\label{eq31}
\end{equation}
where $\epsilon_\theta(x_k,k,\mathbb{O})$ is the output of the denoising network, which estimates the magnitude of the Gaussian noise added to $x_k$ based on the current environment state $\mathbb{O}$, the noisy decision $x_k$, and the time step $k$. The hyperbolic tangent activation function is used to limit the output amplitude, thereby avoiding instability in the decision recovery process caused by excessive prediction noise. 

To enable the gradient of the loss function to back-propagate to the network parameters $\theta$, we apply the reparameterization trick to sample $x_{k-1}$ from the conditional distribution $p_\theta(x_{k-1}\mid x_k)$ as follows~\cite{du2024diffusion}
\begin{equation}
x_{k-1}=\mu_\theta(x_k,k,\mathbb{O})+\sqrt{\tilde{\beta}_k}\epsilon.
\label{eq32}
\end{equation}

By sequentially applying~\eqref{eq32} from $k=K$ down to $k=1$, the reverse process can transform pure noise into the optimal AI agent migration decision $x_0$.

\textit{3) Algorithm Architecture}
\label{s4-2-3}

The framework of CGDM includes an actor network $\pi_\theta$, a double critic network $Q_\phi$, a target actor network $\hat{\pi}_{\hat{\theta}}$, a target critic network $\hat{Q}_{\hat{\phi}}$, an environment, and an experience replay buffer $\mathcal{D}$, as shown in Fig.~\ref{fig3}.

In CGDM, the actor $\pi_\theta$ generates decisions through the reverse process. Specifically, $\pi_\theta$ takes the state $\mathbb{O}$ and the noisy action $\mathbb{A}_K$ as input, and obtains the original action $\mathbb{A}$ after gradual denoising. The reverse denoising mechanism ensures the expressiveness of the multi-modal decision structure and avoids the traditional reliance on unimodal distributions in policy networks. However, if the maximization of the value function $Q_\phi$ is directly used as the optimization goal, the actor may overfit the local high-value area given by the value function, resulting in the degradation of the multi-modal expression of policy~\cite{huang2023reparameterized}. Moreover, the actor could easily deviate from the prior distribution in high-dimensional action spaces, resulting in unstable policy updates. To this end, we propose an adaptive confidence mechanism based on a diffusion model-driven denoising consistency loss, which dynamically adjusts the optimization objective of the actor.

The denoising consistency loss for the actor at a random time step $k\in[1,K]$ is computed as
\begin{equation}
    \mathcal{L}_{\mathrm{BC}}(\theta;\mathbb{O},\mathbb{A})=\mathbb{E}_{k\sim\mathcal{U}\{1,K\}}\|\epsilon_\theta(\mathbb{A}_k,k,\mathbb{O})-\epsilon\|^2,
\label{eq33}
\end{equation}
where $\epsilon \sim \mathcal{N}(0,\mathbf{I})$ is the actual noise added, $\epsilon_\theta$ is the predicted noise by the enoising network $\epsilon_{\theta}$, and $\mathbb{A}_k$ is the noisy action computed from the original action $\mathbb{A}$ according to~\eqref{eq26}. This loss measures the reliability of the denoising process. Thus, we define the confidence of actor as
\begin{equation}
w_a(\mathbb{O},\mathbb{A})
= \exp\!\left(-\kappa\,\mathcal{L}_{\mathrm{BC}}(\theta;\,\mathbb{O},\mathbb{A})\right),
\label{eq34}
\end{equation}
where $\kappa>0$ is a hyperparameter that controls the sensitivity to denoising consistency loss $\mathcal{L}_{\mathrm{BC}}$.

To maximize the action Q-value while maintaining consistency between the noise injection and denoising processes, the confidence in~\eqref{eq34} is used to adaptively regulate the weight of the Q-value. When $\mathcal{L}_{\mathrm{BC}}$ is large, it indicates that the actor has an insufficient grasp of the action prior, causing the confidence $w_a$ to approach $0$. Directly using the Q-value of such actions to update the policy would likely lead to over-exploration and distribution collapse~\cite{li2024learning}. Therefore, the confidence is used to attenuate the influence of the Q-value on the policy network, while the denoising consistency term enhances the reliability of the denoising process, thereby ensuring policy stability. Conversely, when $\mathcal{L}_{\mathrm{BC}}$ is small, the confidence $w_a$ approaches $1$, and the actor should fully leverage Q-value information to guide updates of the policy network toward generating higher-value actions. The optimization objective of the actor is expressed as
\begin{equation}
\begin{aligned}
\max_\theta\;&
\mathbb{E}_{\mathbb{O}\sim\mathcal{D}}\!\left[
    w_a\bigl(\mathbb{O},\pi_\theta(\mathbb{O})\bigr)
    Q_\phi\bigl(\mathbb{O},\pi_\theta(\mathbb{O})\bigr)
\right] \\
&\;-\;
\rho\mathbb{E}_{(\mathbb{O},\mathbb{A})\sim\mathcal{D}}\!\left[
    \mathcal{L}_{\mathrm{BC}}(\theta;\mathbb{O},\mathbb{A})
\right],
\end{aligned}
\label{eq35}
\end{equation}
where the coefficient $\rho$ controls the weight of the behavior clone term. We use the Adam optimizer to optimize the objective function in~\eqref{eq35}, and the gradient of the actor network parameter $\theta$ can be calculated as
\begin{equation}\label{eq36}
\resizebox{0.89\linewidth}{!}{$
\begin{aligned}
\nabla_{\theta}\,\mathcal{L}_{\mathrm{a}}(\theta)
&= \mathbb{E}_{\mathbb{O}\sim\mathcal{D}}\!\Bigl[
    -\nabla_{\theta}\bigl(
        w_a(\mathbb{O},\pi_{\theta}(\mathbb{O}))
        Q_{\phi}(\mathbb{O},\pi_{\theta}(\mathbb{O}))
    \bigr)
\Bigr] \\
&\quad+\;
\mathbb{E}_{(\mathbb{O},\mathbb{A})\sim\mathcal{D}}\!\Bigl[
    \nabla_{\theta}\mathcal{L}_{\mathrm{BC}}(\theta;\mathbb{O},\mathbb{A})
\Bigr].
\end{aligned}
$}
\end{equation}

Subsequently, the actor network parameters are iteratively optimized by gradient descent, and the update speed is controlled by the learning rate $\eta_a$, which is given by
\begin{equation}
    \theta^{\prime}\leftarrow \theta - \eta_a \cdot \nabla_{\theta}\,\mathcal{L}_{\mathrm{a}}(\theta).
\label{eq37}
\end{equation}

To mitigate the overestimation of the value function, we adopt a double critic structure and use Temporal Difference (TD) to optimize the value network parameters $\phi=\{\phi_1,\phi_2\}$.  First, a batch of experiences $(\mathbb{O},\mathbb{A},\mathbb{R},\mathbb{O}')$ is randomly sampled from the replay buffer, and the target actor outputs the next action $\mathbb{A}'=\hat{\pi}_{\hat{\theta}}(\mathbb{O}')$ based on the next state $\mathbb{O}'$. The TD target $\hat{y}$ can be expressed as
\begin{equation}
    \hat{y}=\mathbb{R}+\gamma\min\left\{\hat{Q}_{\hat{\phi}_1}(\mathbb{O}',\mathbb{A}'),\hat{Q}_{\hat{\phi}_2}(\mathbb{O}',\mathbb{A}')\right\},
\label{eq38}
\end{equation}
where $\gamma \in(0,1)$ controls the degree to which future rewards are discounted. The value network parameters are optimized by minimizing the mean squared error, which is expressed as
\begin{equation}
    \operatorname*{min}_{\boldsymbol{\phi}^{1},\boldsymbol{\phi}^{2}}\quad\mathbb{E}_{(\mathbb{O},\mathbb{A},\mathbb{R},\mathbb{O}')\sim\mathcal{D}}[\sum_{i=1,2}\left(\hat{y}-Q_{\phi_i}(\mathbb{O},\mathbb{A})\right)^{2}].
\label{eq39}
\end{equation}

\begin{algorithm}[!t]
\caption{Confidence‑Regulated GDMs}
\begin{algorithmic}[1]
\STATE \textbf{Initialize} actor network $\pi_\theta$, double critic networks $Q_\phi$, target networks $\hat{\pi}_{\hat{\theta}} \gets {\pi}_{\theta}$, $\hat{Q}_{\hat{\phi}} \gets Q_{\phi}$, the environment, and an experience replay buffer $\mathcal{D}$;
\FOR{each training epoch $e=1,2,\ldots,E$}
    \FOR{each trajectory $m=1,2,\ldots,M$}
        \STATE Obtain the current observation $\mathbb{O}$ from the environment;
        \STATE Sample initial noise $x_{T} \sim \mathcal{N}(0,\mathbf{I})$;
        \FOR{denoising step $k = K$ down to $1$}
            \STATE Infer the noise $\epsilon_\theta(x_k,k,\mathbb{O})$ added to $x_k$ using the actor network;
            \STATE Compute the mean of the reverse distribution $p_\theta(x_{k-1}\mid x_k)$ using~\eqref{eq31};
            \STATE Sample $x_{k-1}$ from the reverse process using the reparameterization trick~\eqref{eq32};
        \ENDFOR
        \STATE Process the final output $x_0$ to obtain the action $\mathbb{A}$;
        \STATE Execute the action $\mathbb{A}$ in the environment to receive reward $\mathbb{R}$ and next observation $\mathbb{O}'$;
        \STATE Record the transition $(\mathbb{O}, \mathbb{A}, \mathbb{R}, \mathbb{O}')$ in the replay buffer $\mathcal{D}$;
    \ENDFOR
    \STATE Sample a batch of transitions from $\mathcal{D}$;
    \STATE Compute the denoising consistency loss $\mathcal{L}_{\mathrm{BC}}(\theta;\mathbb{O},\mathbb{A})$ and get the confidence $w_a(\mathbb{O},\mathbb{A})$ of actor, as defined in~\eqref{eq33} and~\eqref{eq34};
    \STATE Update the actor parameters $\theta$ using~\eqref{eq37};
    \STATE Update the critic parameters $\phi$ by minimizing the loss in~\eqref{eq39};
    \STATE Soft-update target network parameters $\hat{\theta},\hat{\phi}$ using~\eqref{eq40};
\ENDFOR
\STATE \textbf{return} the optimized actor network $\pi_{\theta}$;
\end{algorithmic}
\label{alg:CGDM}
\end{algorithm}

To ensure training stability, the target network parameters of both the actor and critic are updated through a soft update mechanism, which are given by
\begin{equation}
\begin{split}
\hat{\theta}' &\leftarrow \tau\,\theta + (1-\tau)\,\hat{\theta},\\
\hat{\phi}'   &\leftarrow \tau\,\phi   + (1-\tau)\,\hat{\phi},
\end{split}
\label{eq40}
\end{equation}
where $\tau \in (0,1)$ is the soft update coefficient used to prevent drastic fluctuations in the parameters during training.

In summary, the CGDM algorithm adaptively regulates the optimization objective in~\eqref{eq35} through the confidence mechanism. It actively leverages value information to optimize the policy when confidence is high and reinforces the structural priors of the denoising process when confidence is low. Consequently, it enables stable exploration and facilitates near‑optimal solutions in high‑dimensional action spaces without expert data. The pseudocode of the CGDM algorithm is presented in Algorithm~\ref{alg:CGDM}.

\textit{4) Complexity Analysis}
\label{s4-2-4}

The computational complexity of the CGDM algorithm consists of two main parts. The first part is the trajectory collection. The algorithm runs for $E$ iterations, and during each iteration, the agent interacts with the environment to collect $M$ trajectories, each with an interaction cost of $F$, and every decision is obtained by a reverse denoising process comprising $K$ steps performed by the actor network. The computational cost of each denoising step is linearly related to the size of the network parameters $|\theta|$, resulting in a trajectory collection complexity of $O(E M (F + K|\theta|))$~\cite{du2024diffusion}. The second part is the network parameter update. In each iteration, a batch of $N$ samples is randomly sampled from the replay buffer to update the actor and double critic networks, with computational complexities of $O(E N |\theta|)$ and $O(E N |\phi|)$~\cite{kang2024uav}, respectively. Additionally, the soft updates of target networks incur a complexity of $O(E(|\theta| + |\phi|))$.  Therefore, the overall computational complexity of the CGDM algorithm is $O(E[M(F+K|\theta|)+(N+1)(|\theta|+|\phi|)])$.

\section{Numerical Results}
\label{s5}

In this section, we comprehensively evaluate the performance of the CGDM algorithm. We first present the experimental configuration, and then we conduct an ablation study on the key components of the CGDM algorithm. We further compare CGDM with several baseline algorithms to analyze its convergence. Finally, we evaluate the performance of CGDM and baseline algorithms under various scenario configurations. 

\subsection{Experimental Settings}
\label{s5-1}
In the simulation experiment, we uniformly deploy four RSUs, which are interconnected through physical links. The vehicles move along their respective driving trajectories, and each vehicle deploys and dynamically migrates the corresponding AI agent to adapt to real-time changes in the vehicle positions and network environment. Vehicles establish wireless communication links with the RSUs within their coverage areas to upload multi-modal data and receive computed results in real time. Additionally, attackers are introduced into the environment and launch intermittent DDoS attacks on specific RSUs, causing temporary service disruptions. The key parameters of the experiment are shown in Table~\ref{tab2}~\cite{kang2024uav,chen2023multiagent,du2024diffusion}.

\subsection{Ablation Study of CGDM Algorithm}
\label{s5-2}

\begin{figure}[!t]
\centering
\includegraphics[width=0.45\textwidth]{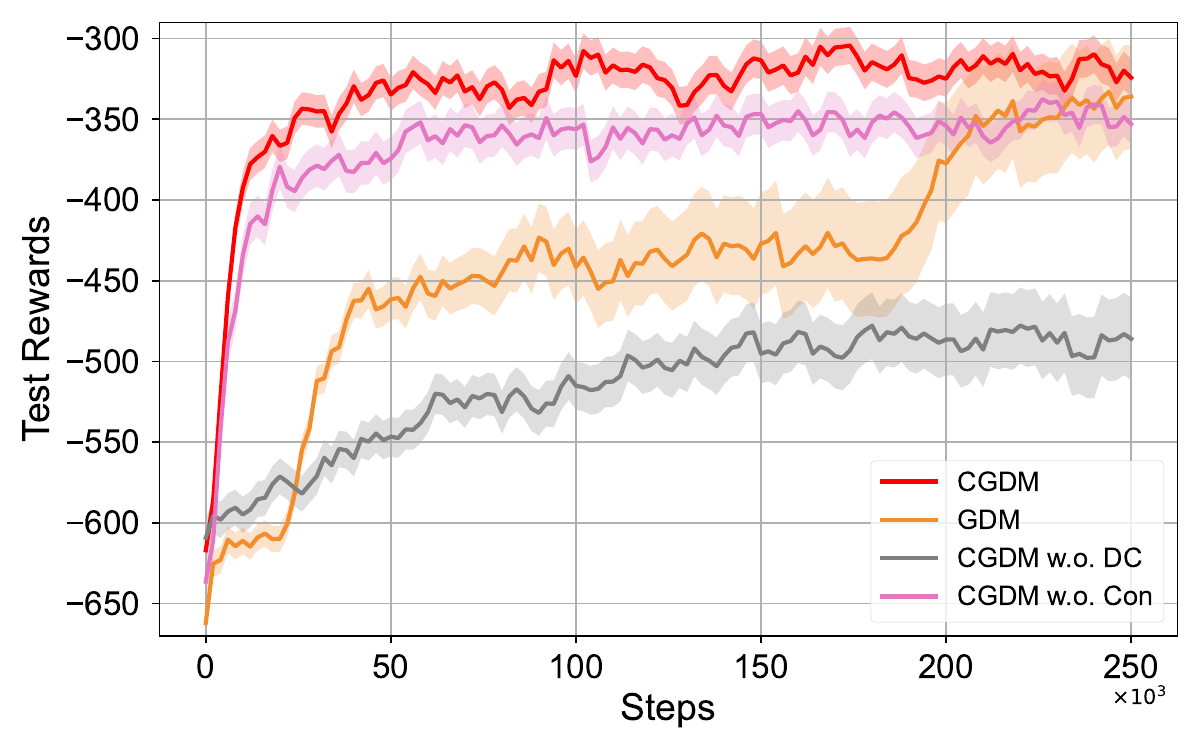}
\caption{Comparison between CGDM with and without confidence mechanism or denoising consistency term.}
\label{ablation}
\end{figure}

In Fig.~\ref{ablation}, we show the convergence performance of the CGDM algorithm when specific components are removed, aiming to explore the role of each component within CGDM. Specifically, we removed the confidence mechanism (i.e., $\kappa=0$) and the denoising consistency term (i.e., $\rho=0$), corresponding to the test reward curves labeled CGDM w.o. Con and CGDM w.o. DC, respectively. We also include the Generative Diffusion Model (GDM) for comparison, which also leverages a diffusion model as a policy and has shown excellent performance in network optimization tasks~\cite{du2024enhancing}.

\begin{table}[!t]
\centering
\caption{Key Parameter Setting}
\label{tab2}
\begin{tabular}{>{\centering\arraybackslash}p{1.1cm}>{\raggedright\arraybackslash}p{3.85cm}>{\centering\arraybackslash}p{1.95cm}} \toprule
 \textbf{Parameters}& \textbf{Descriptions}& \textbf{Values}\\ \midrule
    $V$&Number of vehicles & $8$ \\ 
    $S$&Number of RSUs& $4$ \\ 
    $C_{\mathfrak{Z}}$&Computing capability range of all RSUs&$\text{[$100$, $300$] \rm{MHz}}$\\
    $D_{\mathcal{V}}^{con}$&Size range of multi-modal data by all vehicles& $\text{[$100$, $600$] \rm{MB}}$\\
 $B_{\mathcal{V},\mathcal{S}}$& Wireless bandwidth between vehicles and RSUs&$\text{[$100$, $300$] \rm{Mbps}}$\\
 $\zeta_{\mathcal{A}},\zeta_{\mathcal{S}},\zeta_{\mathcal{P}}$& Weights of each factor of the reputation value&$0.33$\\
    $\xi$& The update speed of reputation values&$0.7$\\ 
    $E$&Total number of episodes& $200$\\
    $K$&  Denoising steps&$5$\\
    $D$&  Maximum capacity of the replay buffer&$1\times10^{6}$\\
     $M$&  Batch size&$256$\\
    $\eta_{\mathrm{a}}$&  Learning rate of the actor  networks&$1\times10^{-4}$\\
    $\eta_{\mathrm{c}}$& Learning rate of the critic networks&$1\times10^{-3}$\\
    $\tau$& Weight of soft update&$0.005$\\
    $\gamma$& Discount factor &$0.95$\\\bottomrule
\end{tabular}
\end{table}

From the test reward curves, we observe that when removing the confidence mechanism, the CGDM w.o. Con scheme still achieves relatively fast convergence and stability compared with the GDM-based DRL algorithm. This occurs due to the persistent regularization effect from the denoising consistency term, which constrains the denoising process and prevents the policy from deviating excessively from the diffusion prior when pursuing high-Q-value actions. However, in the absence of confidence regulation, the final solution remains suboptimal. Conversely, removing the denoising consistency term significantly reduces both convergence speed and decision quality. Without the regularization provided by the denoising consistency term, the policy optimization overly prioritizes maximizing action Q-values. The confidence mechanism alone merely scales the Q-value without effectively preventing the policy from falling into suboptimal solutions. In contrast, the CGDM algorithm combines the denoising consistency term with an adaptive confidence mechanism. This combination enhances policy adherence to the diffusion prior when confidence is low and leverages Q-value guidance when confidence is high. Therefore, the CGDM algorithm ensures stable training process and achieves the best overall performance.

\begin{figure}[!t]
\centering
\includegraphics[width=0.45\textwidth]{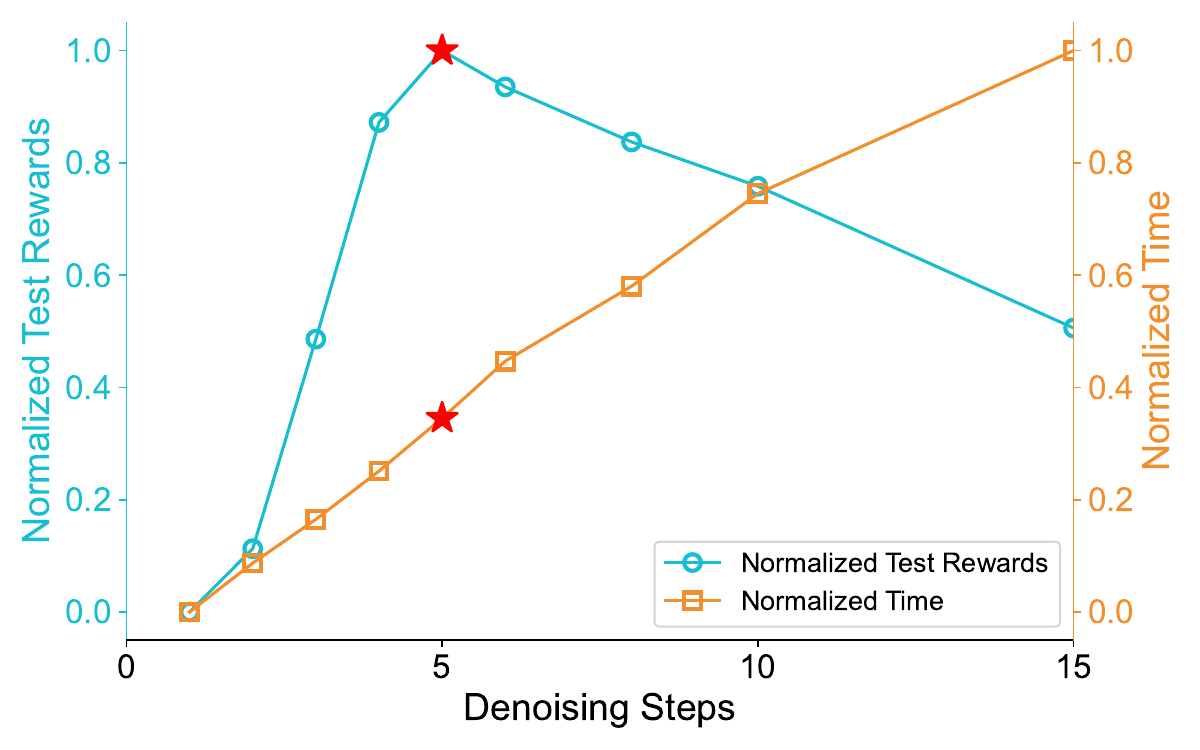}
\caption{Comparison of normalized test rewards and training time with different denoising steps.}
\label{denoising_step}
\end{figure}

Moreover, we analyze the impact of the denoising step $K$ on the performance of CGDM in Fig.~\ref{denoising_step}, selecting  $K\in\{1,2,3,4,5,6,8,10,15\}$ and normalizing both the maximum test reward and training time. Although training time grows linearly with increasing $K$, consistent with theoretical complexity, performance does not improve monotonically. The maximum reward rapidly increases up to $K=5$ and subsequently declines, with a significant performance drop observed after $K=10$. This is because excessive denoising steps lead to gradient explosion and degrade policy stability~\cite{wang2025finetuning}. Consequently, to balance computational efficiency with optimal performance, we select the turning point of the test reward curve, i.e., denoising step $K=5$, as the default setting for subsequent evaluations.

\subsection{Convergence Analysis}
\label{s5-3}

\begin{figure}[!t]
\centering
\includegraphics[width=0.45\textwidth]{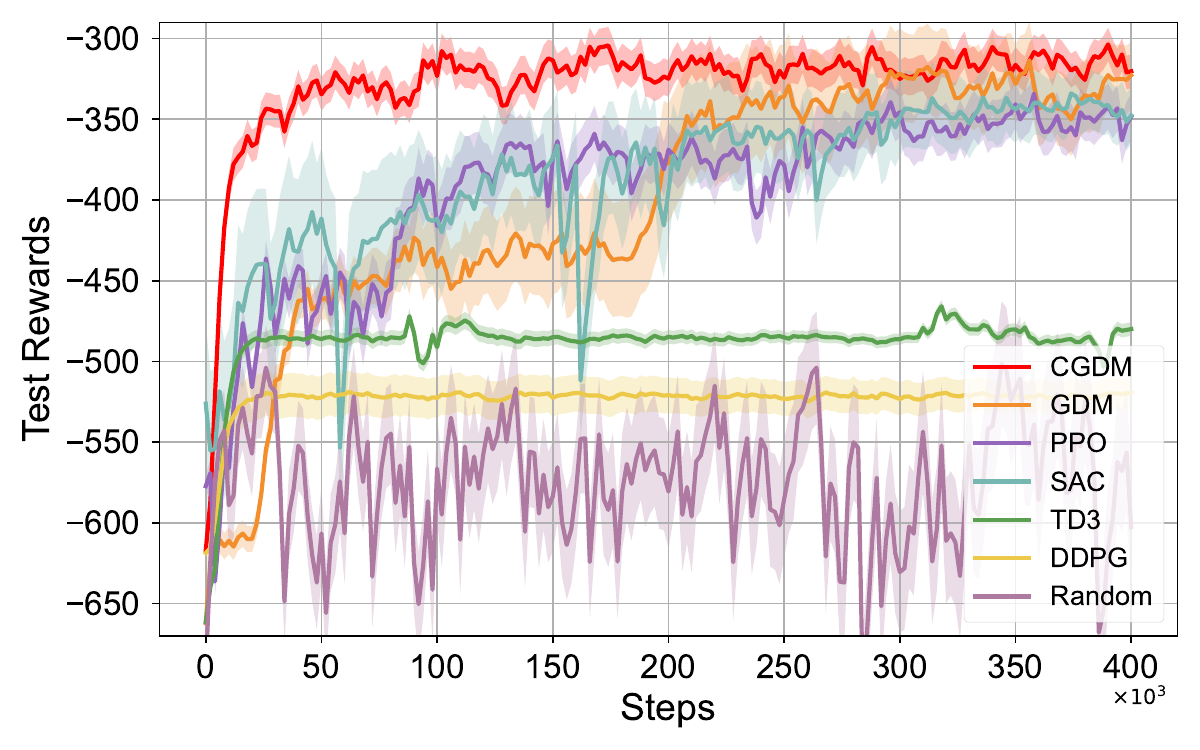}
\caption{Comparison of test reward curves of different algorithms.}
\label{reward}
\end{figure}

As shown in Fig.~\ref{reward}, we compare the training convergence performance of our proposed CGDM algorithm with the GDM algorithm, Proximal Policy Optimization (PPO) algorithm, Soft Actor-Critic (SAC) algorithm, Twin Delayed Deep Deterministic policy gradient (TD3) algorithm, Deep Deterministic policy gradient (DDPG) algorithm, and a Random policy algorithm. From the overall performance, the CGDM algorithm shows superior convergence speed and policy quality, with performance improvements of 5.7$\%$, 12.3$\%$, 10.2$\%$, 34.6$\%$, 39.3$\%$, and 44.8$\%$ compared with GDM, PPO, SAC, TD3, DDPG, and the Random algorithm, respectively.

We can observe that the GDM algorithm exhibits slow performance improvements during the training steps between approximately $5\times10^4$ and $20\times10^4$. This slowdown arises because the optimization objective of GDM solely focuses on maximizing the action Q-value without the constraint provided by denoising consistency loss during the denoising process of the diffusion model. Consequently, the policy tends to become trapped in local regions with relatively high Q-values. Only after approximately $20\times10^4$ training steps does the policy manage to explore beyond these local regions. In contrast, our proposed CGDM algorithm incorporates an adaptive confidence mechanism based on denoising consistency loss, enabling it to autonomously adjust its optimization objective, thereby quickly and stably converging to a global optimum.

Furthermore, CGDM significantly outperforms classical DRL algorithms. Although the PPO algorithm demonstrates high training stability, its policy distribution is limited to a unimodal Gaussian form, making it challenging to capture the multi-modal characteristics inherent to vehicular AI agent migration decisions in high-dimensional action spaces. Similarly, the SAC algorithm, despite promoting exploration by maximizing policy entropy, also employs a unimodal Gaussian distribution to represent policies, inevitably resulting in convergence to suboptimal solutions. Deterministic policy gradient methods such as TD3 and DDPG exhibit inadequate exploration in decision space, causing them to easily converge to poor-performing local regions in high-dimensional action scenarios. Compared with the above algorithms, CGDM effectively captures the multi-modal characteristics of policy distribution through the denoising mechanism of diffusion models, significantly enhancing decision quality and stability in complex environments.

\subsection{Performance Evaluation}
\label{s5-4}

\begin{figure}[!t]
\centering
\includegraphics[width=0.45\textwidth]{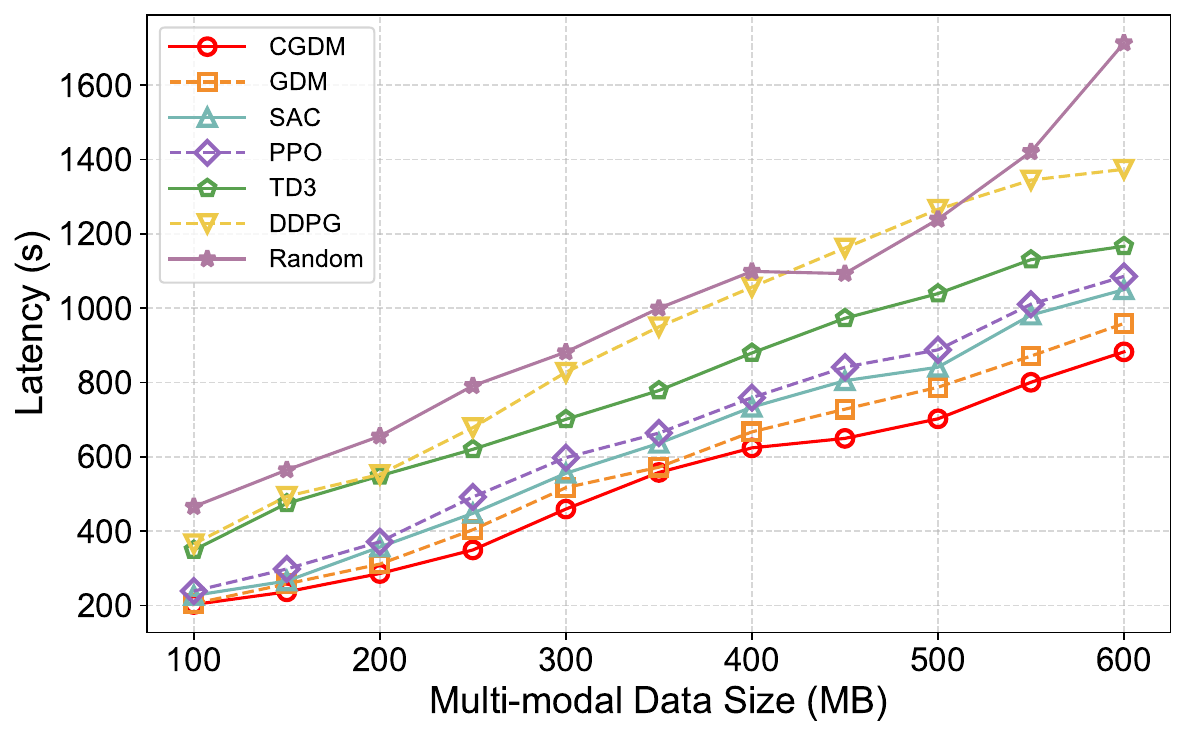}
\caption{Total system latency under different sizes of raw multi-modal data uploaded by vehicles.}
\label{data_size}
\end{figure}

To demonstrate the robustness of the CGDM algorithm, we analyze its performance under various system parameters.

In Fig.~\ref{data_size}, we illustrate the change in the total system latency as the size of the multimodal data uploaded by vehicles increases from $100$ \rm{MB} to $600$ \rm{MB}. The results indicate that the CGDM algorithm consistently achieves the lowest system latency as the data size grows. Compared with GDM, SAC, PPO, TD3, DDPG, and Random algorithms, CGDM reduces latency by 7.6$\%$, 16.3$\%$, 19.9$\%$, 35.6$\%$, 43.8$\%$, and 48.7$\%$, respectively. When data volumes are small, performance differences among algorithms are similar. However, as data volume increases, performance gaps become more pronounced. This occurs because larger data sizes significantly raise the upload and processing latency between vehicles and RSUs. Owing to its diffusion-based decision approach, the CGDM algorithm effectively optimizes RSU resource allocation, thereby reducing the total system latency and ensuring an immersive user experience.

\begin{figure}[!t]
\centering
\includegraphics[width=0.45\textwidth]{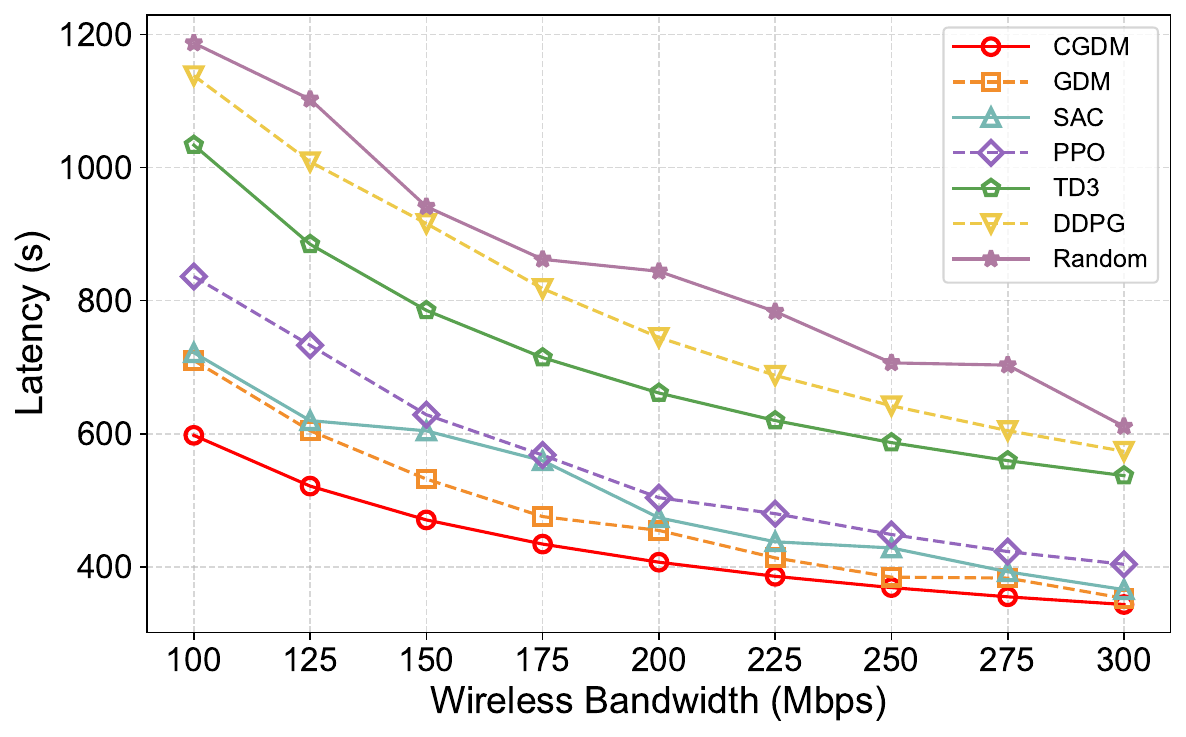}
\caption{Total system latency under different wireless bandwidths.}
\label{wireless_bandwidth}
\end{figure}

\begin{figure}[!t]
\centering
\includegraphics[width=0.45\textwidth]{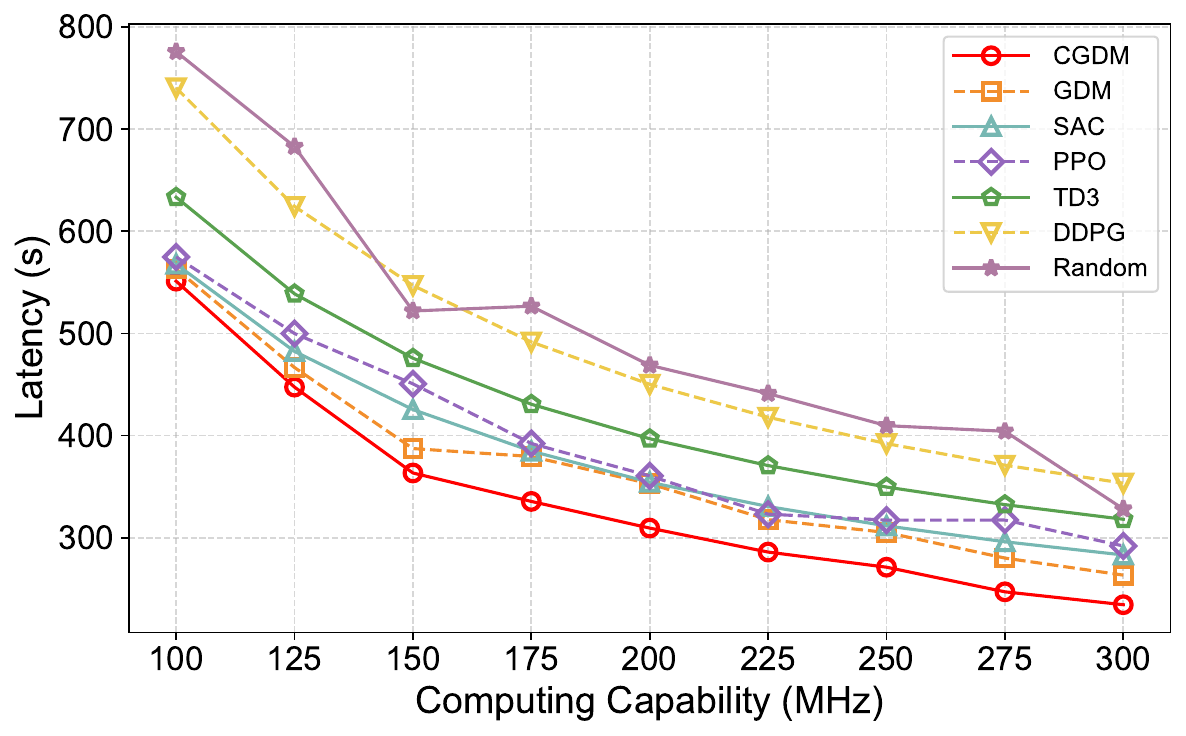}
\caption{Total system latency under different computing capabilities of RSUs.}
\label{com_cap}
\end{figure}

In Fig.~\ref{wireless_bandwidth}, we show the optimization effects of different algorithms as wireless bandwidth increases from $100$ \rm{Mbps} to $300$ \rm{Mbps}. Under limited bandwidth conditions, CGDM efficiently allocates RSU's computing resources, reducing overall system latency. As bandwidth increases, wireless transmission latencies decrease rapidly. At this stage, total system latency is dominated by task processing latency at RSUs. When all algorithms are constrained by the same computing resource of RSUs, the latency gap among them tends to decrease. Specifically, CGDM reduces system latency by 9.0$\%$, 14.8$\%$, 21.5$\%$, 38.7$\%$, 44.9$\%$, and 49.5$\%$ compared to GDM, SAC, PPO, TD3, DDPG, and Random algorithms, respectively.

In Fig.~\ref{com_cap}, we present the system performance as the computing capacity of the RSU increases from $100$ \rm{MHz} to $300$ \rm{MHz}. With the increase in computing capacity, the performance gap among the algorithms gradually widens. However, when the computing capacity is further enhanced beyond $250$ \rm{MHz}, the differences narrow again. According to the system model, the computing capacity directly influences data processing latency. With moderate computing resources, the CGDM algorithm can allocate resources efficiently, avoiding both redundant and insufficient resource allocation among RSUs, thus showcasing significant performance advantages. When each RSU has surplus computing capacity, the performance gaps among the algorithms decrease. Numerical results demonstrate that CGDM reduces system latency by 8.9$\%$, 12.3$\%$, 14.5$\%$, 21.7$\%$, 31.1$\%$, and 33.4$\%$ compared with GDM, SAC, PPO, TD3, DDPG, and Random algorithms, respectively.

\begin{figure}[!t]
\centering
\includegraphics[width=0.45\textwidth]{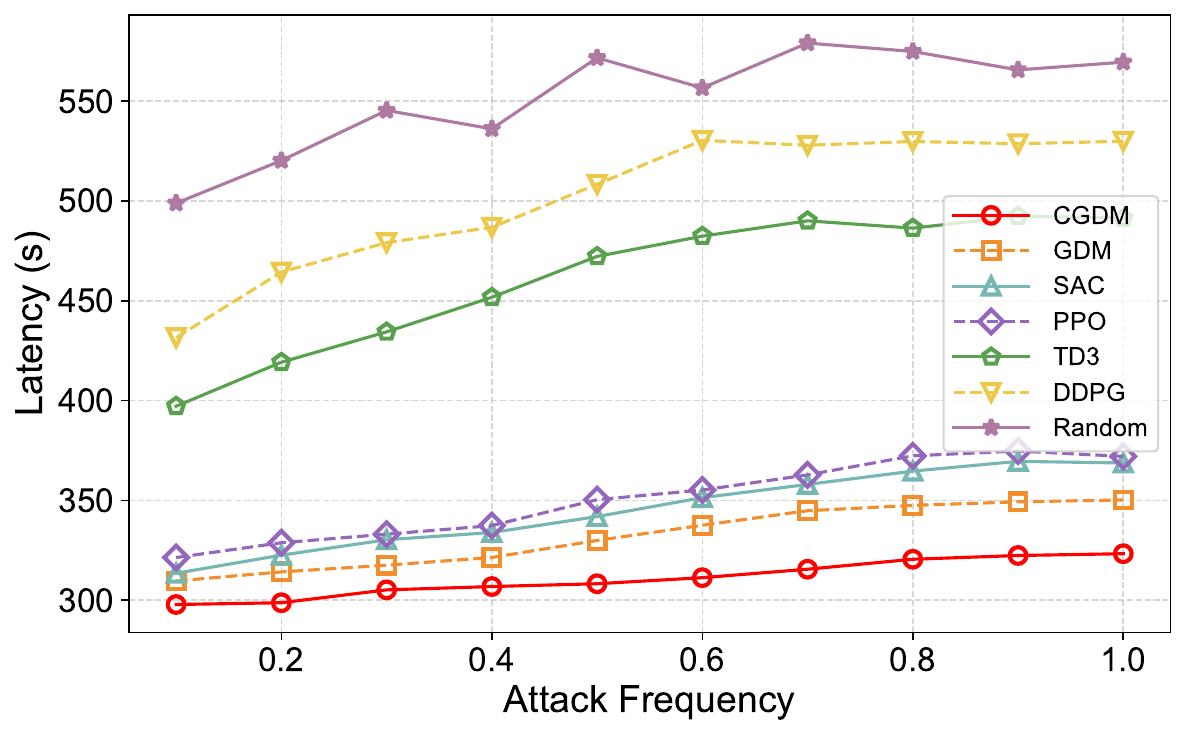}
\caption{Total system latency under different attack frequencies.}
\label{attack_fre}
\end{figure}

\begin{figure}[!t]
\centering
\includegraphics[width=0.45\textwidth]{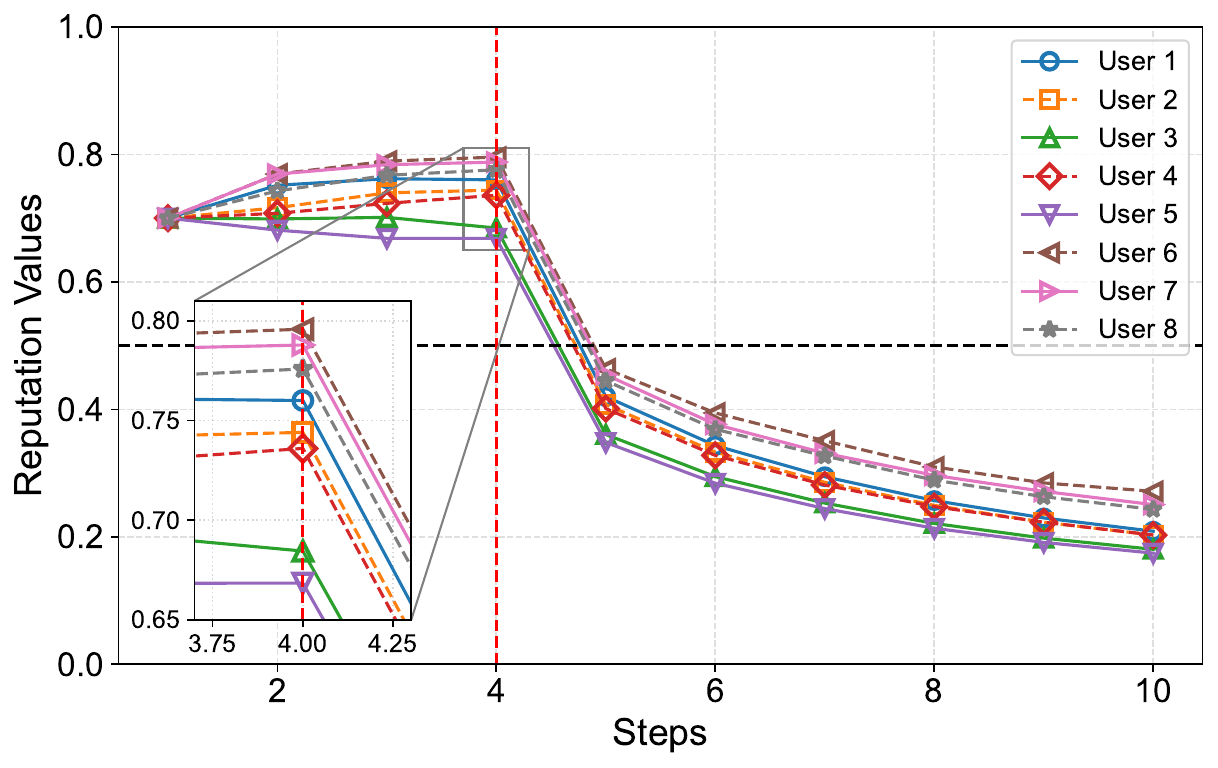}
\caption{Reputation values of RSUs for different users in one episode.}
\label{repu_change}
\end{figure}

In Fig.~\ref{attack_fre}, we show the impact of varying attack frequencies on the system latency. The CGDM algorithm consistently demonstrates the lowest latency across different attack frequencies, achieving latency reductions of 6.3$\%$, 9.8$\%$, 11.2$\%$, 32.4$\%$, 37.8$\%$, and 43.6$\%$ compared with GDM, SAC, PPO, TD3, DDPG, and Random algorithms, respectively. At lower attack frequencies, performance differences among algorithms are relatively minor. However, as the attack frequency exceeds $0.5$, the CGDM algorithm demonstrates a significant advantage, reflecting its stronger capability to resist attacks and control latency. This is because CGDM can accurately identify the RSU under attack based on changes in RSU reputation values, and dynamically optimize the migration decision, thereby effectively avoiding the performance degradation caused by high-frequency attacks. Moreover, as the attack intensity reaches saturation, the latency curves of each algorithm tend to be flat, limiting further impact on system latency from increasing attack frequencies.

Additionally, to evaluate the effectiveness of the TPB-based trust evaluation model, we record the reputation value changes of RSUs towards different users within one episode, as shown in Fig.~\ref{repu_change}. We set the initial reputation value of RSU for all users to $0.7$. After several steps of interaction, the reputation values exhibit obvious personalized differences, reflecting the ability of the TPB-based evaluation model to effectively capture user trust preferences. When the attack occurs, the reputation value of RSUs for each user quickly drops below the safety threshold, indicating that the TPB-based trust evaluation model can accurately and timely identify RSUs under attack, thereby effectively ensuring the security of vehicular AI agent migration.

\section{Conclusion}
\label{s6}

In this paper, we have studied the real-time migration of AI agents in the vehicular metaverse. To address potential security threats and trust biases during migration, we have proposed a TPB-based trust evaluation model, which objectively quantifies the reputation of RSUs while fully considering personalized user trust preferences. Additionally, we have formulated the vehicular AI agent migration optimization problem as a POMDP and introduced a CGDM algorithm to solve it. The proposed CGDM algorithm leverages diffusion models to effectively capture multi-modal decision distributions and employs a denoising consistency term and an adaptive confidence mechanism to ensure stable and efficient policy updates. Numerical results have demonstrated that our method achieves superior convergence speed and robustness compared to multiple baseline algorithms. For future work, we plan to extend the CGDM algorithm to multi-agent collaborative scenarios to generate decisions more efficiently.

\bibliographystyle{IEEEtran}
\bibliography{ref}

\end{document}